\documentclass[5p,final,authoryear,twocolumn]{elsarticle}

\usepackage{amssymb}
\usepackage{graphicx}
\usepackage{amsmath}
\usepackage{subfigure}
\usepackage[latin1]{inputenc}
\usepackage{color}
\usepackage{multirow}
\usepackage[hypcap=false]{caption}
\usepackage{rotating}
\usepackage{url}

\DeclareMathOperator*{\argmax}{arg\,max}

\DeclareRobustCommand\onedot{\futurelet\@let@token\@onedot}
\def\@onedot{\ifx\@let@token.\else.\null\fi\xspace}

\definecolor{gold}{rgb}{0.85,.66,0}   

\newcommand{\revision}[1]{\textcolor{black}{#1}}

\graphicspath{{pics/}{figs/}{figs/segOverlays/}{../Miccai/figs/}{../Miccai/}{figs/figuresRev/}{figs/figures/}}

\usepackage{booktabs}
\journal{NeuroImage}

\begin{document}
\begin{frontmatter}

\title{QuickNAT: A Fully Convolutional Network for Quick and Accurate Segmentation of Neuroanatomy}

\author{Abhijit Guha Roy$^{a,b}$ \corref{corradd}}
\cortext[corradd]{{Corresponding Author. Address: Waltherstr. 23, 80337 M\"unchen, Germany; Email: abhi4ssj@gmail.com}}
\author{Sailesh Conjeti$^{b,c}$, Nassir Navab$^{b,d}$, Christian Wachinger$^{a}$}

\address{$^{a}$Artificial Intelligence in Medical Imaging (AI-Med), Department of Child and Adolescent Psychiatry, LMU M\"{u}nchen, Germany \\
$^{b}$Computer Aided Medical Procedures, Department of Informatics, Technical University of Munich, Germany \\
$^{c}$German Center for Neurodegenerative Diseases (DZNE), Bonn, Germany \\
$^{d}$Computer Aided Medical Procedures, Johns Hopkins University, Baltimore, USA.
}

\begin{abstract}

Whole brain segmentation from structural magnetic resonance imaging (MRI) is a prerequisite for most morphological analyses, but is computationally intense and can therefore delay the availability of image markers after scan acquisition. We introduce QuickNAT, a fully convolutional, densely connected neural network that segments a \revision{MRI brain scan} in 20 seconds. To enable training of the complex network with millions of learnable parameters using limited annotated data, we propose to first pre-train on auxiliary labels created from existing segmentation software. Subsequently, the pre-trained model is fine-tuned on manual labels to rectify errors in auxiliary labels. With this learning strategy, we are able to use large neuroimaging repositories without manual annotations for training. In an extensive set of evaluations on eight datasets that cover a wide age range, pathology, and different scanners, we demonstrate that QuickNAT achieves superior segmentation accuracy and reliability in comparison to state-of-the-art methods, while being orders of magnitude faster. The speed up  facilitates processing of large data repositories and supports translation of imaging biomarkers by making them available within seconds for fast clinical decision making. 
\end{abstract}


\end{frontmatter}

\section{Introduction}
Magnetic Resonance Imaging (MRI) provides detailed in-vivo insights about the morphology of the human brain, which is essential for studying development, aging, and disease~\citep{giedd_brain_1999, draganski_neuroplasticity:_2004, shaw_intellectual_2006, raznahan_prenatal_2012, alexander-bloch_imaging_2013, wachinger_whole-brain_2016, lerch_studying_2017}.
In order to access measurements like volume, thickness, or shape of a structure, the neuroanatomy needs to be segmented, which is a time-consuming process when performed manually~\citep{fischl_freesurfer}. 
Computational tools have been developed that can fully automatically segment brain MRI scans by warping a manually segmented atlas to the target scan~\citep{fischl_freesurfer,ashburner_unified_2005,rohlfing_quo_2005,svarer_mr-based_2005}. 
Such approaches have two potential shortcomings: (i) the estimation of the 3D deformation field for warping is computationally intense, and (ii) lack of homologies may result in erroneous segmentations of the cortex~\citep{lerch_studying_2017}. Due to these drawbacks, existing atlas-based methods require hours of processing time for each scan and may result in sub-optimal solutions.  

We propose a method for the \emph{Quick segmentation of NeuroAnaTomy} (QuickNAT) in MRI T1 scans based on a deep fully convolutional neural network (F-CNN) that runs in seconds on GPUs, compared to hours for existing atlas-based methods. 
We believe that this speed up by several orders of magnitude can have a wide impact on neuroimaging: processing of large datasets can be performed on a single GPU workstation, instead of a computing cluster; quantitative morphological measurements can be derived from a scan within seconds, boosting its translation. 
Furthermore, the fast processing speed allows for sampling multiple segmentations in a reasonable amount of time to estimate segmentation uncertainty for automated quality control~\citep{roy2018inherent}. 
Beside its speed, QuickNAT produces state-of-the-art segmentation accuracy as demonstrated on multiple datasets covering a wide age range, different field strengths, and  pathologies. Moreover, it yields effect sizes that are closer to those of manual segmentations and therefore offers advantages for group analyses. Finally, QuickNAT exhibits high test-retest accuracy making it useful for longitudinal studies. 

Deep learning models have had ample success over the last years, but require vast amounts of annotated data for effective training~\citep{lecun_deep_2015}. The task of semantic image segmentation is dominated by F-CNN models in computer vision~\citep{long_fully_2015}. The limited availability of training data with manual annotations presents the main challenge in extending F-CNN models to brain segmentation. To address this challenge, we introduce a new training strategy (Fig.~\ref{fig:training}) that exploits large brain repositories without manual labels and small repositories with manual labels. First, we apply existing software tools (e.g., FreeSurfer~\citep{fischl_freesurfer}) to segment scans without annotations. We refer to these automatically generated segmentations as \emph{auxiliary labels}, which we use to pre-train the network. Auxiliary labels may not be as accurate as expert annotations; however, they allow us to efficiently leverage the vast amount of initially unlabeled data for supervised training of the network. It also makes the network familiar with a wide range of morphological variations of different brain structures that may exist in a wide population. In the second step, we fine-tune (i.e., continue training) the previous network with smaller manually annotated data. Pre-training provides a good prior initialization of the network, such that scarce manual annotations are optimally utilized to achieve high segmentation accuracy.  
As a side note, we observed that a network trained only on FreeSurfer segmentations can produce more accurate results than FreeSurfer itself. 

QuickNAT consists of three 2D F-CNNs operating on coronal, axial and sagittal views followed by a view aggregation step to infer the final segmentation (Fig.~\ref{fig:viewAgg}). Each F-CNN has the same architecture and is inspired by the traditional encoder/decoder based U-Net architecture with skip connections~\citep{u-net}, enhanced with unpooling layers~\citep{deconvnet} (Fig.~\ref{fig:arch}). We also introduce dense connections~\citep{densenet} within each encoder/decoder block to aid gradient flow and to promote feature re-usability, which is essential given the limited amount of training data. The network is optimized using a joint loss function of multi-class Dice loss and weighted logistic loss, where weights compensate for high class imbalance in the data and encourage proper estimation of anatomical boundaries. 

The two main methodological innovations of QuickNAT are the training strategy with auxiliary labels and the F-CNN architecture. To the best of our knowledge, this is the first work to conduct such a large number of experiments on highly heterogeneous datasets to evaluate the robustness of an F-CNN for brain segmentation. The code and trained model are available as extensions of MatConvNet~\citep{vedaldi_matconvnet:_2015} at \url{https://github.com/abhi4ssj/QuickNATv2}. This is an extension of our early work \citep{roy_error_2017}, where we introduced the concept of pre-training with auxiliary labels. In this work, we improved upon the architecture, segment more brain structures and show exhaustive experiments for a wide range of possibilities to substantiate the effectiveness of the framework.

\section{Methods}
Given an input MRI brain scan $I$, we want to infer its segmentation map $S$, which indicates $27$ cortical and subcortical structures. Given a set of scans $\mathcal{I} = \{I_1, \dots I_n \}$ and its corresponding segmentations $\mathcal{S} = \{ S_1, \dots, S_n \}$, we want to learn a function $f_{seg}:I \rightarrow S$. We express this function as an F-CNN model, termed QuickNAT, which is detailed below.

\subsection{Architectural Design}
QuickNAT has an encoder/decoder like 2D F-CNN architecture with $4$ encoders and $4$ decoders separated by a bottleneck layer shown in Fig.~\ref{fig:arch}. The final layer is a classifier block with softmax. The architecture includes skip connections between all encoder and decoder blocks of the same spatial resolution, similar to the U-Net architecture~\citep{u-net}. These skip connections not only provide \revision{encoder feature information} to the decoders, but also provide a path of gradient flow from the shallower layers to deeper layers, improving training. In the decoder stages, instead of up-sampling the feature maps by convolution transpose like U-Net, we included un-pooling layers~\citep{deconvnet}. These ensure appropriate spatial mappings of the activation maps during up-sampling, which in turn improves segmentation accuracy, especially for small subcortical structures. 

\subsubsection{Dense Block}
\begin{figure*}[t]
\centering
\includegraphics[width=0.65\textwidth]{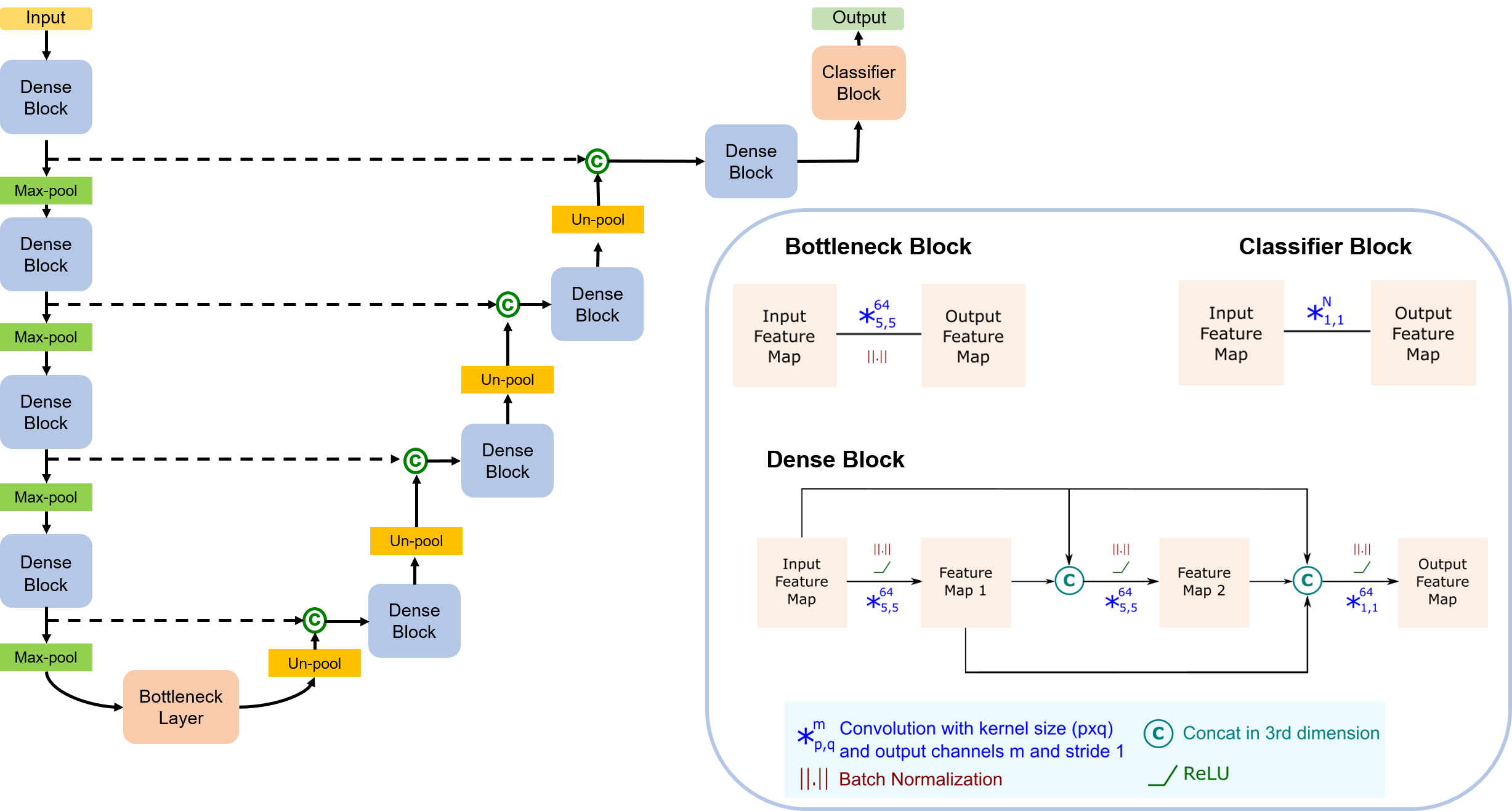}
\caption{Illustration of QuickNAT's encoder-decoder based fully convolutional architecture consisting of dense, bottleneck and classifier blocks shown in the zoomed view. The symbols corresponding to different network layers are also explained here.}
\label{fig:arch}
\end{figure*}

Each dense block consists of three convolutional layers (Fig.~\ref{fig:arch}). Every convolutional layer is preceded by a batch-normalization layer and a Rectifier Linear Unit (ReLU) layer. The first two convolutional layers are followed by a concatenation layer that concatenates the input feature map with outputs of the current and previous convolutional blocks. These connections are referred to as dense connections~\citep{densenet} which improves gradient flow during training and promote feature re-usability across different stages of convolution~\citep{densenet}. In addition, they help learning better representations promoting features learned by different convoltional layers within the same block to be different. The kernel size for these two convolutional layers is kept small, $5\times5$, to limit the number of parameters. Appropriate padding is provided so that the size of feature maps before and after convolution remains constant. The output channels for each convolution layer is set to $64$, which acts as a bottleneck for feature map selectivity. The input channel size is variable, depending on the number of dense connections. The third convolutional layer is also preceded by a batch normalization and ReLU, but has a $1\times1$ kernel size to compress the feature map size to $64$. A flow diagram of the dense block is illustrated in Fig.~\ref{fig:arch}.

\subsubsection{Encoding Path}
The encoder path consists of a set of $4$ dense blocks, each followed by a $2\times2$ max-pooling block, which at each stage reduces the spatial dimension of the feature maps by half. During down-sampling by max-pooling, the indices corresponding to the maximum activations are saved and passed to decoder blocks for un-pooling.

\subsubsection{Bottleneck}
The bottleneck block consists of a $5\times5$ convolutional layer and a batch normalization layer to separate the encoder and decoder part of the network, restricting information flow between the encoder and decoder.

\subsubsection{Decoding Path}
The decoder path also consists of $4$ dense blocks. Each dense block is preceded by an un-pooling layer. This layer recovers the actual spatial locations corresponding to the maximum activations, which are lost during max-pooling in the encoders, and places them at the correct location during up-sampling~\citep{deconvnet}. This is very relevant when segmenting small subcortical structures. Another, advantage of up-sampling is that it does not require any learnable parameters in comparison to convolutional transpose used in U-Net~\citep{u-net}. The up-sampling is followed by a skip-connection, which concatenates the un-pooled feature map with the output feature map of the corresponding encoder before max-pooling. Skip connections add \revision{encoder features to the decoders for aiding segmentation} and thus allow gradients to flow from deeper to shallower regions of the network. The concatenated feature map is passed to the next dense block with similar architecture.

\subsubsection{Classifier Block}
The output feature map from the last decoder block is passed to the classifier block, which is basically a convolutional layer with $1\times1$ kernel size that maps the input to an $N$ channel feature map, where $N$ is the number of classes ($28$ in our case). This is followed by a softmax layer to map the activations to probabilities, so that all the channels represent probability maps for each of the classes.

\subsection{Loss Function}
We train QuickNAT by optimizing two loss functions simultaneously: (i) the weighted logistic loss, and (ii) the multi-class Dice loss. The logistic loss provides a pixel-wise probabilistic estimate of similarity between the estimated labels and the manually annotated labels. The Dice loss is inspired from the Dice overlap ratio, which estimates similarity between the estimated and manually annotated labels~\citep{v-net}. It was initially introduced for two-class segmentation and  we extend it to multi-class segmentation in this work. Given the estimated probability $p_l(\mathbf{x})$ of pixel $\mathbf{x}$ belonging to class $l$ and the its actual class $g_l(\mathbf{x})$, the loss function is 

\begin{equation}
\mathcal{L} =  \underbrace{ -\sum_{\mathbf{x}} \omega(\mathbf{x}) g_l(\mathbf{x}) \log(p_{l}(\mathbf{x}))}_{\mathrm{Logistic Loss}} - \underbrace{\frac{2 \sum_{\mathbf{x}} p_{l}(\mathbf{x}) g_{l}(\mathbf{x})}{\sum_{\mathbf{x}} p_{l}^2(\mathbf{x}) + \sum_{\mathbf{x}} g_{l}^2(\mathbf{x})}}_{\mathrm{Dice Loss}}.
\label{eq:cost}
\end{equation}
 
The first term is the multi-class logistic loss and the second term is the Dice loss. We introduce weights $\omega(\mathbf{x})$, which balance the relative importance of pixels in the loss. We use weights to address two challenges: (i) class imbalance, and (ii) errors in segmentations at anatomical boundaries.
Given the frequency $f_l$ of class $l$ in the training data, i.e., the class prior probability, the indicator function $\mathbb{I}$, the training segmentation~$S$, and the 2D gradient operator $\nabla$, the weights are defined as
\begin{equation}
\omega(\textbf{x}) = \sum_l I( S(\mathbf{x})=l) \ \frac{median(\mathbf{f})}{f_l} + \omega_0 \cdot \mathbb{I}(|\nabla S(\mathbf{x})|>0)
\label{eq:weight}
\end{equation}
with the vector of all frequencies $\mathbf{f} = [f_1, \ldots, f_N]$.
The first term models median frequency balancing~\citep{segnet} and compensates for the class imbalance problem by up-weighting rare classes in the image. 
The second term puts higher weight on anatomical boundary regions to encourage correct segmentation of contours. $\omega_0$ is set to $\frac{2 \cdot median(f)}{f_{min}}$ to give higher priority to boundaries.

\subsection{Model Learning}
We train QuickNAT with stochastic gradient descent with momentum. 
The learning rate is chosen such that proper convergence on validation data is achieved.
It is initially set to $0.1$ and reduced by one order after every $10$ epochs during pre-training.
The training is conducted \revision{until the validation loss plateaus}. 
We use a constant weight decay of $0.0001$. Batch size is set to $4$, limited by the 12GB RAM of the NVIDIA TITAN X Pascal GPU. Momentum is set to a high value of $0.95$ to compensate for noisy gradients due to a small batch size. 
Our choice for the weight decay constant and momentum is based on settings for other modern CNNs.
Prior to inputing scans into the network, they are processed with `\texttt{mri-convert --conform}' from the FreeSurfer pipeline~\citep{fischl_freesurfer}, which performs basic standardization and runs in about 1 second.


\begin{figure*}[h]
\centering
\includegraphics[width=0.7\textwidth]{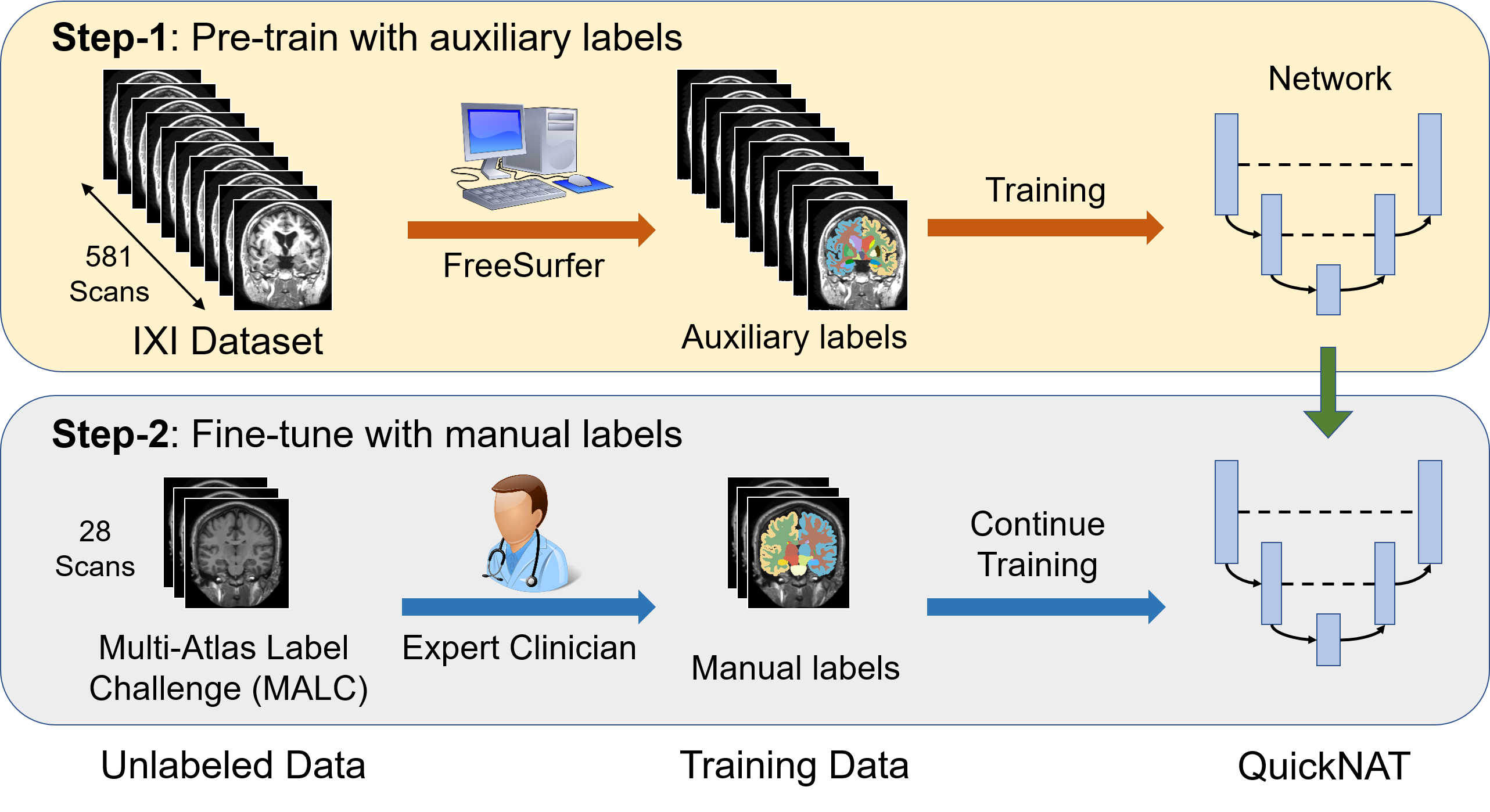}
\caption{Illustration of the two-step training strategy for QuickNAT. First, we use an existing segmentation software (e.g., FreeSurfer) to automatically segment a large unlabeled corpus (IXI Dataset with $581$ scans). These labels are referred to as auxiliary labels and used to pre-train QuickNAT. Second, we fine-tune the network on $28$ scans from the multi-atlas label challenge, which were manually annotated by an expert. Fine-tuning does not start from scratch, but continues optimizing the pre-trained model from step 1, to maximally benefit from the scarce data with manual annotations.}
\label{fig:training}
\end{figure*}

\subsection{Training with Limited Annotated Data}
F-CNN models directly produce a segmentation for all image pixels in an end-to-end fashion without splitting the image into patches. Therfore, they can fully exploit the image context, avoid artificial partitioning of an image, and achieves an enormous speed-up. Yet, training F-CNNs is challenging because each image serves as a single training sample and consequently much larger datasets with manual labels are required than for patch-based approaches, where each image yields many patches. While the amount of unlabeled data rapidly grows, the access to labeled data is still limited due to the labor intense process of manual annotations. At the same time, the success of deep learning is mainly driven by supervised learning, while unsupervised approaches are still an active field of research. Data augmentation artificially increases the training dataset by simulating different variations of the same data, but it cannot encompass all possible anatomical variations observable in a population. We propose to process unlabeled data with existing software tools to create auxiliary labels. These auxiliary labels are not optimal; however, they allow us to use the vast amount of initially unlabeled data for supervised pre-training of the network, enforcing a strong prior. The training procedure consists of two main steps (Fig.~\ref{fig:training}):
\begin{enumerate}
\item \textbf{Pre-training on large unlabeled datasets with auxiliary labels:}  In this step, we use a large neuroimaging dataset (IXI dataset~\footnote{http://brain-development.org/ixi-dataset/}) and process it with an existing tool to create auxiliary labels. The IXI dataset was acquired from three centers and is characterized by a high age range of participants and substantial anatomic variability. We apply the widely used FreeSurfer~\citep{fischl_freesurfer} to obtain auxiliary segmentations, but other tools could be used, depending on the application. We pre-train QuickNAT on this large dataset with auxiliary labels, which results in a network that imitates FreeSurfer segmentations. Pre-training enforces a strong prior on the network, where robustness to data heterogeneity is encouraged by the diversity of the IXI dataset.

\item \textbf{Fine-tuning with limited manually labelled data:}  In this step, we take the pre-trained model and fine-tune it with small data having manual annotations. Instead of learning all filters from scratch, fine-tuning only focuses on rectifying discrepancies between auxiliary and manual labels. We use data from the Multi-Atlas Labelling Challenge dataset~\citep{malc} for fine-tuning. During this, we lower the initial learning rate to $0.01$ and reduce it by an order of magnitude after every $5$ epochs until convergence.
\end{enumerate}

\subsection{Multi-View Aggregation}
\begin{figure*}[t]
\centering
\includegraphics[width=0.65\textwidth]{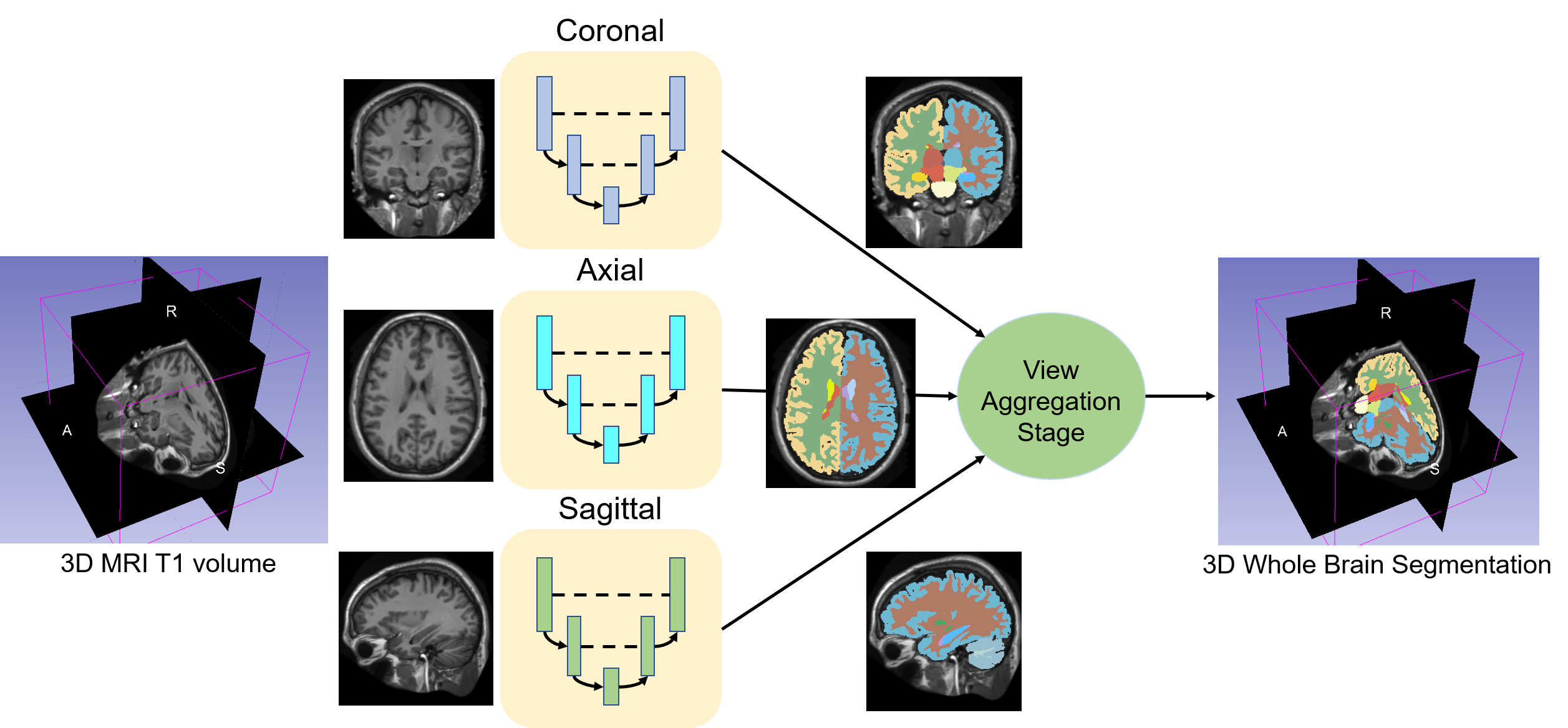}
\caption{We show the multi-view aggregation step that combines segmentations from models trained on 2D slices along three principal axes: coronal, sagittal and axial. The final segmentation is obtained by combining the probability maps from all the three networks.}
\label{fig:viewAgg}
\end{figure*}

To also consider the third dimension in QuickNAT, we train a separate F-CNN for each of the three principal views: coronal, axial and sagittal (Fig.~\ref{fig:viewAgg}). The predictions for each these networks are combined into the final segmentation in a multi-view aggregation step. The final label for a voxel $\mathbf{x}$ is given by $L_{Pred}(\mathbf{x})$, which is computed as

\begin{equation}
L_{Pred}(\mathbf{x}) = \argmax_c ( \lambda_1 p_{Ax}(\mathbf{x}) + \lambda_2 p_{Cor}(\mathbf{x}) +\lambda_3 p_{Sag}(\mathbf{x}) )
\end{equation}

\noindent
where $p_{Ax}(\mathbf{x})$, $p_{Cor}(\mathbf{x})$, $p_{Sag}(\mathbf{x})$ are the predicted probability vectors for axial, coronal and sagittal views respectively, and $\lambda_1$, $\lambda_2$, $\lambda_3$ their corresponding fixed weights. The probability score for a particular structure reflects the certainty of the network in the prediction, which depends on how well the structure is represented in the corresponding view. Aggregating all the votes for a voxel $\mathbf{x}$ provides a regularization effect for the label prediction and reduces spurious predictions. 

The highly symmetric layout of the brain poses challenges for segmentation in sagittal slices, as it is not possible to differentiate slices from the left and right hemisphere. Thus, we assign structures from the left and right hemisphere the same label number for training on sagittal slices. This reduces the number of classes from 28 to 16. At testing, we re-map the probability maps from 16 to 28 structures by replicating probabilities for left and right. Due to this, we assign a lower value to $\lambda_3$ in comparison to $\lambda_1$ and $\lambda_2$. In our case, we set $\lambda_1,\lambda_2,\lambda_3$ to 0.4, 0.4 and 0.2, respectively, to give relatively equal importance to all views.  

\section{Experimental Datasets}

We use nine brain MRI datasets in our experiments. We use five datasets with manual annotations to evaluate segmentation accuracy. Three datasets were used for testing reliability of the segmentation framework. Table~\ref{tab:dataset} summarizes the number of subjects per dataset, the age range, the diagnosis, and the annotated structures. Present diagnoses are Alzheimer's disease (AD),  mild cognitive impairment (MCI), and psychiatric disorders. Details about acquisition protocol used in each of the datasets can be found in their respective references. All MRI datasets  are publicly available.

\noindent
(i) \textbf{IXI dataset:} The dataset is collected from 3 different hospitals from London (Hammersmith hospital, Guy's hospital, institute of psychiatry) and consists of both 1.5T and 3.0T MRI T1 scans for 581 healthy subjects. The data was collected by Imperial College London and is available for download at \url{http://brain-development.org/ixi-dataset/}.

\noindent
(ii) \textbf{Multi-Atlas Labelling Challenge (MALC):} The dataset is part of the OASIS dataset~\citep{marcus_open_2007} and contains MRI T1 scans from 30 subjects with manual annotations for the whole brain~\citep{malc}. In the challenge, 15 volumes were defined for training and 15 for testing. We follow the same setup in our experiments. The dataset also includes follow-up scans from 5 subjects to evaluate inter-run consistency. Manual segmentations were provided by Neuromorphometrics, Inc. under academic subscription.

\noindent
(iii) \textbf{ADNI-29:}  The dataset is a subset of 29 subjects from the Alzheimer's Disease Neuroimaging Initiative~\citep{jack_tracking_2013} (ADNI, \url{adni.loni.usc.edu}). The primary goal of ADNI has been to test whether serial MRI, PET, other biological markers, and clinical and neuropsychological assessment can be combined to measure progression of MCI and AD. Manual whole-brain segmentations were provided by Neuromorphometrics, Inc. under academic subscription. The dataset includes 15 controls and 14 Alzheimer's patients. Furthermore, 15 scans were acquired with 1.5T field strength and 14 scans with 3.0T, with balanced distribution of AD and controls. 

\noindent
(iv) \textbf{Internet Brain Segmentation Repository (IBSR):} The dataset consists of 18 T1 MRI scans with manual segmentations of the whole brain~\citep{rohlfing2012image}. The MR scans and their manual segmentations were provided by the Center for Morphometric Analysis at Massachusetts General Hospital and are available at \url{http://www.cma.mgh.harvard.edu/ibsr/}. 

\noindent
(v) \textbf{Child and Adolescent NeuroDevelopment Initiative (CANDI):} The dataset consists of 13 T1 MRI scans (8 male, 5 female) of children with psychiatric disorders, having minimum age of 5, maximum age of 15 and mean age of 10 years. Manual whole-brain segmentations were provided by Neuromorphometrics, Inc. under academic subscription. The dataset is publicly available at \cite{kennedy_candishare:_2012}.

\noindent
(vi) \textbf{Harmonized Protocol (HarP):} The European Alzheimer's Disease Consortium and ADNI~\citep{jack_tracking_2013} together provided a Harmonized Protocol (HarP) for manual hippocampal segmentation from MRI \citep{boccardi_training_2015}. It was defined by through an evidence-based Delphi panel that converged on a consensus definition. After standardization, a dataset with 131 volumes was released with manually annotated right/left hippocampus for development and evaluation of standard automated hippocampus segmentation algorithms. Special attention was paid for ensuring that the dataset is representative of physiological and pathological variability (age, dementia severity), field strength (1.5T and 3T) and scanner manufacturer (GE, Philips and Siemens). 45 scans were from AD subjects, 42 scans from Controls and 44 scans from MCI.

\noindent
(vii) \textbf{ALVIN:} \cite{kempton_comprehensive_2011}~released the ALVIN dataset in an attempt to standardize the evaluation of segmentation algorithms. The dataset consists of 7 young adult subjects and 9 subjects with Alzheimer's disease. The dataset does not provide manual segmentations but volume measurements of the ventricles. These volume measurements are available from a manual rater at two points in time to observe intra-observer variability.

\noindent
(viii) \textbf{Test-Retest (TRT) dataset:} This dataset was released to analyze reliability of segmentation frameworks for estimating volumes of brain structures~\citep{maclaren_reliability_2014}. The dataset consists of 120 MRI T1 scans from 3 subjects (40 scans per subject) in 20 sessions (2 scans per session) over the duration of 31 days. All the subjects were healthy aged 26-31 years.

\noindent
(ix) \textbf{Travelling Human Phantom (THP) dataset:} This dataset was released to check the reliability of automated segmentation frameworks in estimating volumes from scans acquired from different sites~\citep{magnotta_multicenter_2012}. In the study, 3 healthy subjects were scanned at 8 different centers in the USA using scanners from different vendors. All scans were acquired within a period of 30 days. The sites are: (1) Cleveland Clinic, (2) Dartmouth, (3) University of Iowa, (4) Johns Hopkins, (5) Massachusetts General Hospital, (6) University of California Irvine, (7) University of Minnesota, (8) University of Washington.

In our experiments, we use FreeSurfer annotations from IXI for pre-training and manual annotations from MALC (provided by Neuromorphomrtrics Inc. under academic license) for fine-tuning. The definitions of anatomical structures between MALC and FreeSurfer are identical. Annotators at Neuromorphometrics Inc. follow the CMA (Center for Morphometric Analysis) protocol, which was also used in the creation of the FreeSurfer atlas.
\revision{ADNI-29 and HarP are a subset from the Alzheimer's Disease
Neuroimaging Initiative (ADNI) database (adni.loni.usc.edu). The ADNI was launched in
2003 as a public-private partnership, led by Principal Investigator Michael W. Weiner,
MD. 
The primary goal of ADNI has been to test whether serial magnetic resonance imaging
(MRI), positron emission tomography (PET), other biological markers, and clinical and
neuropsychological assessment can be combined to measure the progression of mild
cognitive impairment (MCI) and early Alzheimer's disease (AD). 
For up-to-date information,
see www.adni-info.org.}

\begin{table*}[t]
\centering
\caption{Summary of the datasets used for training and testing. Dataset characteristics are shown together with available manually annotated structures. Information regarding the diagnosis of IBSR dataset and Age information of THP dataset were not available.}
  \begin{tabular}{|p{1in}|c|p{1.3in}|c|c|}
    \hline
     \textbf{Dataset} & No. of Subjects & \revision{Age} & Diagnosis & Annotations\\
    \hline
    \textbf{IXI} & 581 &	$49.09\pm16.43$ &	Normal &	None \\ \hline
	\textbf{MALC} &	30 &	$3416\pm20.40$ &	CN/AD/MCI &	Whole Brain \\ \hline
\textbf{ADNI-29} &	29 &	$75.87\pm5.86$ &	CN/AD &	Whole Brain \\ \hline
\textbf{IBSR} &	18 &	$29.05\pm4.80$ &	- &	Whole Brain \\ \hline
\textbf{CANDI} &	13 &	$10.00\pm3.13$ &	Psychiatric Disorders &	Whole Brain \\ \hline
\textbf{ALVIN} &	16 &	AD:~$77.4\pm2.4$, Young adults:~$23.8\pm4.1$ &	CN/AD &	Lateral Ventricle \\ \hline
\textbf{HarP} &	131 &	AD:~$74.2\pm7.8$, CN:~$76.2\pm7.4$, MCI:~$74.7\pm8.1$ &	CN/MCI/AD &	Hippocampus \\ \hline
\textbf{TRT} &	3 (40 scans/subject) &	26-31 &	Normal &	None \\ \hline
\textbf{THP} &	5 (Scans from 8 sites in USA) &	- &	Normal &	None \\ \hline

  \end{tabular}
  \label{tab:dataset}
\end{table*}


\begin{table*}[t]
\centering
\caption{Summary of the experiments for evaluating segmentation accuracy (1-5) and reliability (6-8). For each experiment, the table indicates the datasets used for pre-training, fine-tuning and testing together with the number of scans in parentheses. We also list the purpose for each experiment.}
  \begin{tabular}{|p{1in}|c|c|c|c|p{1.8in}|}
    \hline
    \textbf{Evaluation} & \textbf{Experiment} & \textbf{Pre-training} & \textbf{Fine-tuning} & \textbf{Testing} & \textbf{Purpose of Experiment}\\ \hline 
    \multirow{5}{*}{}{Segmentation Accuracy}
    & Experiment 1 & IXI (581) & MALC (15) & MALC (15) & Benchmark Challenge Dataset. \\ \cline{2-6}
    & Experiment 2 & IXI (581) & MALC (28) & ADNI (29) & Robustness to pathology, scanner field strength, group analysis by effect sizes. \\ \cline{2-6}
    & Experiment 3 & IXI (581) & MALC (28) & IBSR (18) & Robustness to low resolution data with wide age range. \\ \cline{2-6}
    & Experiment 4 & IXI (581) & MALC (28) & CANDI (13) & Robustness to children with psychological disorders. \\ \cline{2-6}
    & Experiment 5 & IXI (581) & MALC (28) & HarP (131) & Robustness of hippocampus segmentation in presence of dementia.  \\ \hline
    \multirow{3}{*}{}{Segmentation Reliability}
    & Experiment 6 & IXI (581) & MALC (28) & ALVIN (16) & Reliable volume estimation of lateral ventricles for young and aged subjects (dementia). \\ \cline{2-6}
    & Experiment 7 & IXI (581) & MALC (28) & TRT (120) & Inter- and intra-session reliability. \\ \cline{2-6}
    & Experiment 8 & IXI (581) & MALC (28) & HTP (67) & Reliability across 8 centers with different scanners. \\   
    \hline
  \end{tabular}
  \label{tab:experiments}
\end{table*}

\section{Experiments and Results}
We evaluate QuickNAT in a comprehensive series of eight experiments to assess accuracy, reproducibility, and sensitivity on a large variety of neuroimaging datasets, summarized in Tab.~\ref{tab:experiments}.
In all experiments, we pre-train QuickNAT on 581 MRI volumes from the IXI dataset to get auxiliary segmentations from FreeSurfer~\citep{fischl_freesurfer}. 
We conducted 5 experiments to evaluate the segmentation accuracy (experiments 1 to 5; Sec.~\ref{sec:samedata} and Sec.~\ref{sec:diffdata}), and another 3 experiments (experiments 6 to 8; Sec.~\ref{sec:segreg}) to assess the reliability and consistency of QuickNAT segmentations.
We divided Experiments 1-5 into two sets, (i) Training and testing on the same dataset (Sec.~\ref{sec:samedata}), and (ii) Training and testing on different dataset, i.e. cross dataset experiments (Sec.~\ref{sec:diffdata}).

\subsection{Evaluation of segmentation accuracy with training and testing on same dataset}
\label{sec:samedata}

\subsubsection{Experiment 1: MALC}

In the first experiment, we use the MALC data and replicate the setup of the original challenge~\citep{malc}. A problem associated with segmenting this dataset is that all the training volumes are from young adults while testing volumes include subjects with 70 years and older (maximum 90 years). To achieve good performance, the network therefore has to be robust to differences due to age.

In this experiment, we compare the performance of QuickNAT with state-of-the-art methods and evaluate the impact of pre-training. 
Tab.~\ref{tab:malc} reports the results measured in Dice overlap score. 
First, we compare with the existing F-CNN models (FCN~\citep{long_fully_2015}, U-Net~\citep{u-net}). 
Along each column, we observe that for all the F-CNN models, pre-training with auxiliary labels followed by fine-tuning (termed `Fine-tuned' in Tab.~\ref{tab:malc}) yields significantly ($p<0.001$) better performance than training only with limited manually annotated data (termed `Only Manual' in Tab.~\ref{tab:malc}). Second, when comparing across rows, we observe that QuickNAT performs better than U-Net and FCN in every setting.
QuickNAT significantly ($p<0.001$) outperforms U-Net and FCN by a margin of $5\%$ points and $12\%$ points mean Dice score after fine-tuning. Noteworthy is that QuickNAT not only is better when trained on the FreeSurfer labels (pre-trained model) but also when trained exclusively with limited manually annotated data.

\begin{table*}[t]
\centering
\caption{Comparison of Dice scores of QuickNAT with state-of-the-art methods on 15 testing scans from MALC. Results of only using the pre-trained model on test data is referred as `\textbf{Pre-Trained}'. Results for training only with 15 manual data from scratch is referred as `\textbf{Only Manual}'.  Results of using the pre-trained model and fine-tune it with 15 manual data is referred as `\textbf{Fine-tuned}'.}
  \begin{tabular}{|p{1.5in}|p{0.9in}|p{0.9in}|p{0.9in}|}
    \hline
     \textbf{Method} & \textbf{Pre-Trained} & \textbf{Only Manual} & \textbf{Fine-Tuned}\\
    \hline
    \textbf{QuickNAT} & $0.798\pm0.097$ & $0.874\pm0.067$ & $\mathbf{0.901\pm0.045}$ \\ \hline
    \textbf{U-Net} & $0.681\pm0.193$ & $0.762\pm0.124$ & $0.857\pm0.079$ \\ \hline
    \textbf{FCN} & $0.579\pm0.245$ & $0.534\pm0.311$ & $0.778\pm0.121$ \\ \hline
    \textbf{DeepNAT} & \multicolumn{3}{c|}{$0.891$ for $25$ structures} \\ \hline
    \textbf{Spatial Staple} & \multicolumn{3}{c|}{$0.879\pm0.063$} \\ \hline
    \textbf{PICSL} & \multicolumn{3}{c|}{$0.898\pm0.050$} \\
    \hline
  \end{tabular}
  \label{tab:malc}
\end{table*}

Next, we compare the fine-tuned QuickNAT model with state-of-the-art atlas-based methods PICSL~\citep{wang_multi-atlas_2013} (winner of challenge) and Spatial STAPLE~\citep{asman_formulating_2012} (top 5 in challenge), and state-of-the-art 3D CNN based deep learning method DeepNAT~\citep{wachinger_deepnat:_2017}. Our model outperforms Spatial STAPLE by a statistically significant margin ($p<0.05$). 
The 15 scans for training and the 15 scans for testing were consistent with the challenge definition, for a fair comparison.
It outperforms PICSL by a small margin, which is not statistically significant. It also outperforms DeepNAT when comparing segmentations for only 25 structures. DeepNAT operates in 3D, however, on patches extracted from the image, which limits the context for prediction. A direct extension to a 3D full volume prediction, instead of patches, is limited by the current available amount of GPU memory. 

\begin{figure*}[t]
\centering
\includegraphics[width=0.999\textwidth]{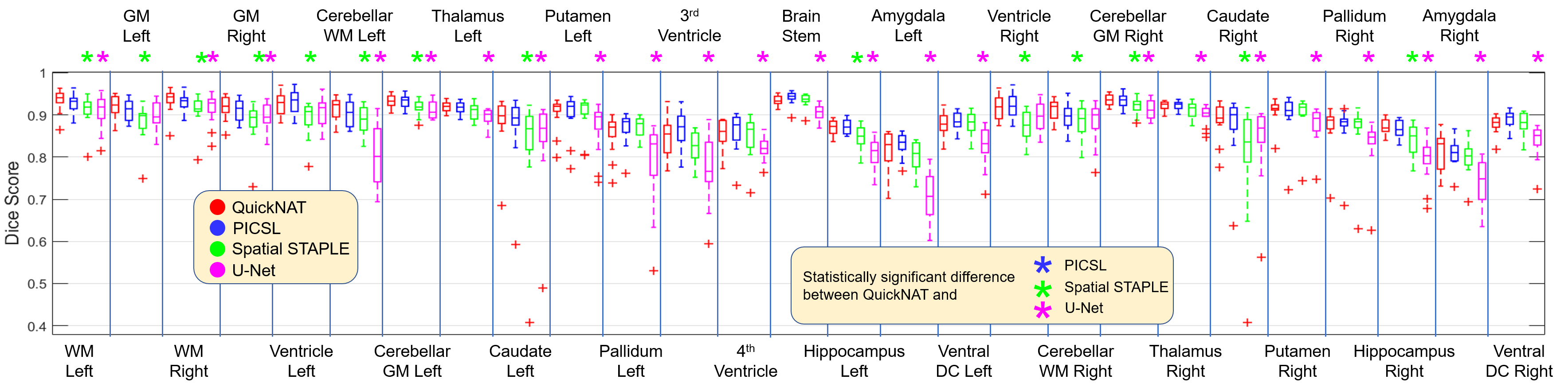}
\caption{Box-plot of Dice scores comparing QuickNAT with PICSL, Spatial STAPLE and U-Net on 15 test volumes of MALC dataset for all the 27 structures. Statistical significance ($p < 0.05$) in comparison to QuickNAT is indicated by a star symbol ($\star$). The p-values were estimates using two-sided Wilcoxon rank-sum test. WM indicates White Matter and GM indicates Grey Matter.}
\label{fig:malc_box}
\end{figure*}

In addition to comparing of brain-wide Dice scores, we performed a structure-wise comparison (see Fig.~\ref{fig:malc_box}). Significant differences are highlighted with a star symbol ($\star$). No significant differences exist across any of the 27 structures to the challenge winner PICSL. QuickNAT has significantly higher Dice for many structures compared to Spatial STAPLE (14 of 27) and U-Net (23 of 27). 

\begin{figure*}[t]
\centering
\includegraphics[width=0.8\textwidth]{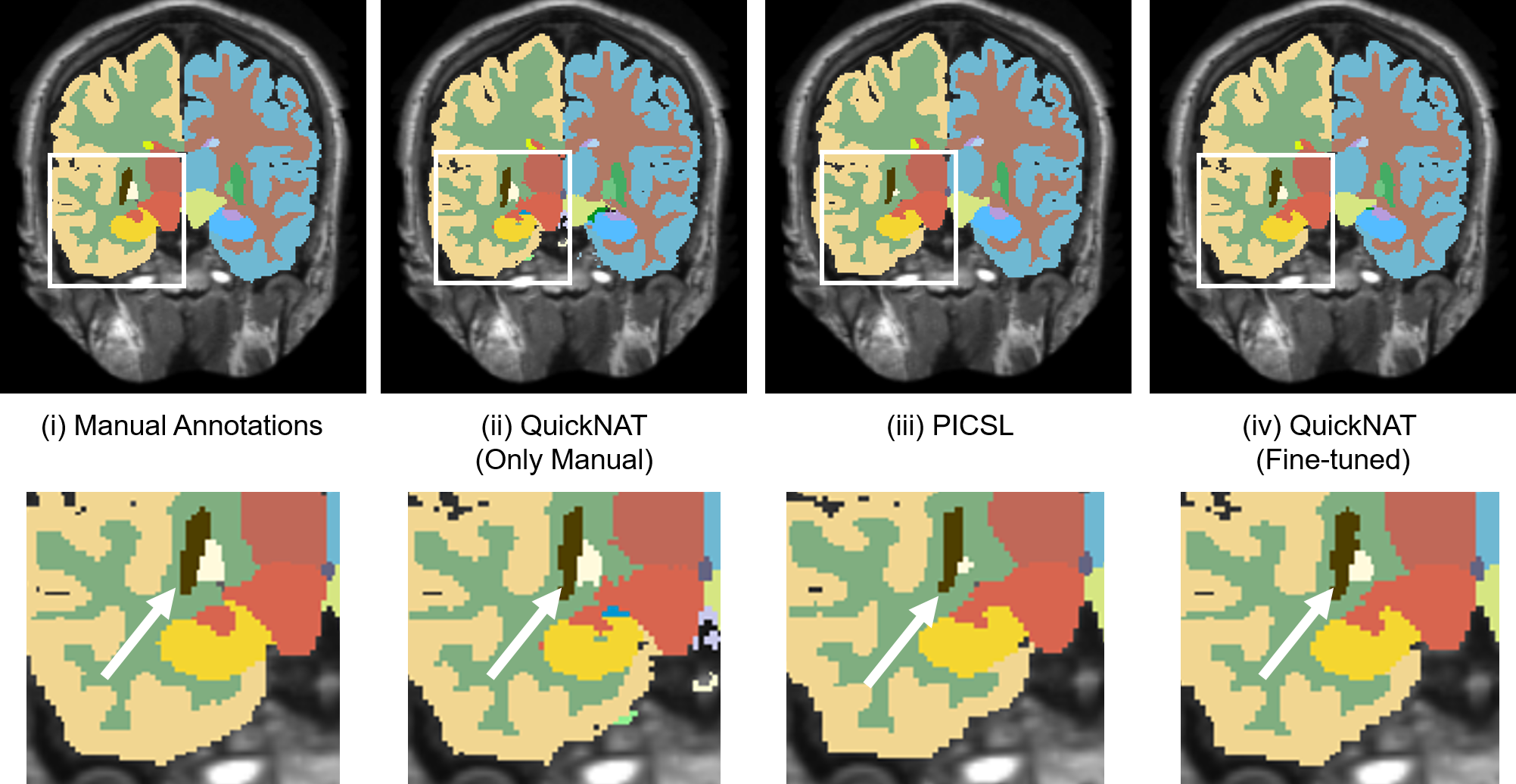}
\caption{Qualitative results of QuickNAT trained with and without pre-training along with PICSL. A zoomed view shows left Putamen (dark brown) and left Pallidum (white) with a white arrow for all the cases indicating the superior segmentation performance of fine-tuned QuickNAT over others.}
\label{fig:malc_seg}
\end{figure*}

\noindent
\textbf{Qualitative Analysis: }
Sample segmentations are visualized in Fig.~\ref{fig:malc_seg} for QuickNAT (trained on small data), PICSL and QuickNAT (fine-tuned) along with the manual segmentation. A zoomed view of the segmentations is also presented. We indicate two important subcortical structures left putamen (brown) and left pallidum (white) with a white arrow. We can observe under-inclusion of left putamen for PICSL. We also observe many spurious misclassified regions in the background in QuickNAT (trained with limited data), which are absent for QuickNAT (fine-tuned).

\begin{figure}[h]
\centering
\includegraphics[width=0.49\textwidth]{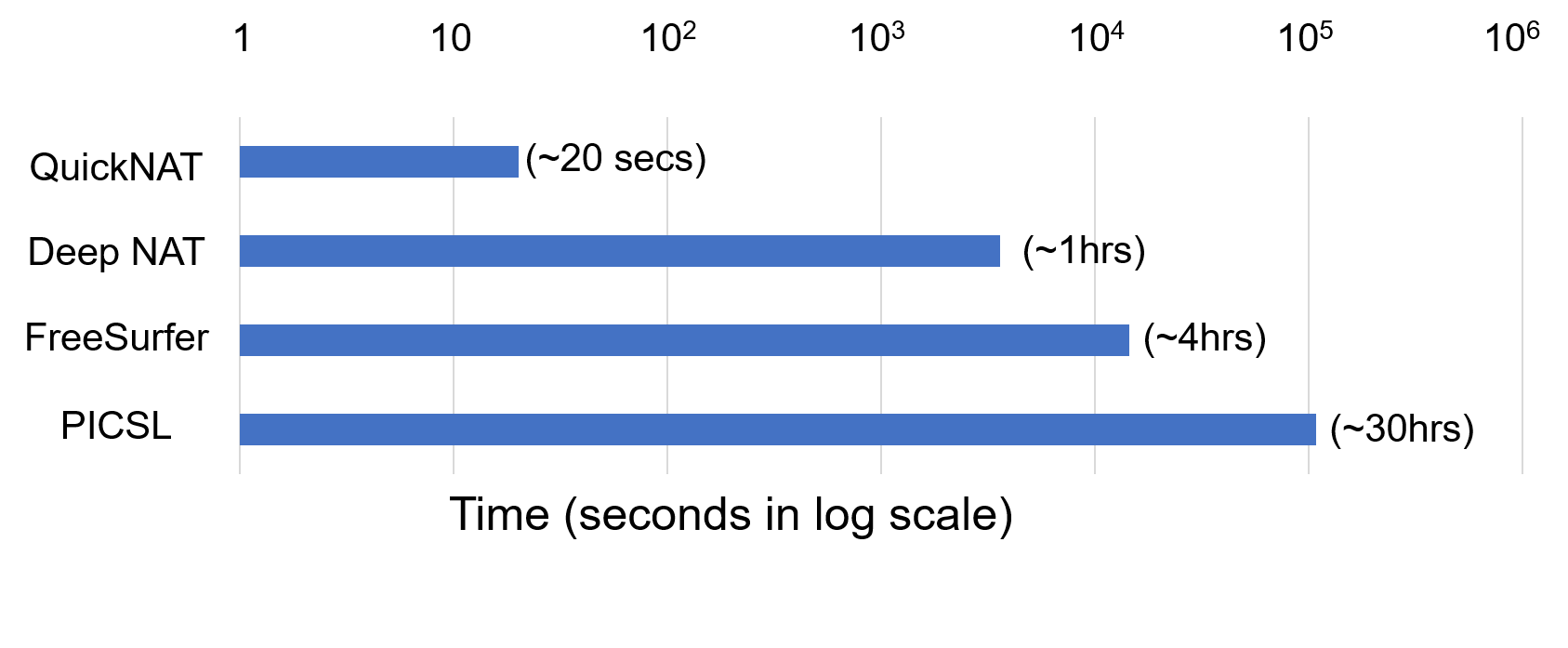}
\caption{illustration of segmentation speed for QuickNAT, DeepNAT, FreeSurfer and PICSL in log scale, demonstrating its superior speed.}
\label{fig:speed}
\end{figure}

\noindent
\textbf{Speed: }
Existing state-of-the-art atlas-based brain segmentation frameworks build upon 3D deformable volume registration, e.g., ANTs~\citep{avants2ducibl011reproe}. 
In pair-wise registrations, each image in the atlas is transformed to the test image. This results in long runtimes, since a single pair-wise registration takes about 2 hours on a 2GHz CPU machine~\citep{malc}. On MALC with 15 training images, the approximate segmentation time for both PICSL and Spatial STAPLE is 30h/vol. FreeSurfer has its own atlas and takes around 4h/vol. DeepNAT uses a 3D patch-based approach for segmentation, which takes around 1h/vol. In comparison to these models, QuickNAT segments a volume in 20s, which is orders of magnitude faster. We illustrate the segmentation time in Fig.~\ref{fig:speed} in logarithmic scale. The speed of QuickNAT can be further reduced to about 6s, if only one anatomical view instead of all three are used for segmentation. We observed the best segmentation performance on a single view for coronal view, with an overall Dice of $0.895\pm0.055$ compared to $0.901\pm0.045$ with view aggregation. 

\noindent
\textbf{Inter-Run Consistency: }
MALC includes a second MRI T1 scan for 5 patients to evaluate the consistency of the segmentation algorithm across acquisitions (test/retest). For quantification, we estimate the volumes of different structures for both runs and compute the volume distance ($d_V$) between them. 
\revision{This metric indicates the error in volume estimation after segmentation. It is defined as
\begin{equation}
d_V(V_a, V_e) = 2\frac{|V_a - V_e|}{V_a + V_e}
\end{equation}
where $V_a$ is the actual volume (estimated from manual segmentation)
and $V_e$ is the estimated volume, for a given structure. Higher volume distance
indicates poor estimation of volume and therefore an indirect measure of low segmentation quality.}
Table~\ref{tab:malc_consis} reports mean volume distance for whole brain, hippocampus, amygdala, lateral ventricles, white matter and grey matter; comparing QuickNAT with PICSL and Spatial STAPLE. In case of whole brain, all values are very low with low standard deviation, indicating that each of the methods produces highly consistent results. For structures hippocampus, white matter, grey matter QuickNAT has a lower volume distance indicating its superiority over PICSL and Spatial STAPLE. For structures Lateral ventricle and amygdala, QuickNAT has poorer performance than PICSL and Spatial STAPLE, but within an acceptable range of $2-3\%$.

\begin{table}[t]
\footnotesize
\centering
\caption{Inter-Run Consistency reported by volume distance over 5 subjects from MALC Dataset chosen by the Challenge organizers.}
  \begin{tabular}{|p{0.8in}|p{0.65in}|p{0.65in}|p{0.65in}|}
    \hline
     \textbf{Structures} & \textbf{Spatial STAPLE} & \textbf{PICSL} & \textbf{QuickNAT}\\ \hline
     Whole Brain & $0.013\pm0.015$ & $0.018\pm0.018$ & $0.017\pm0.016$ \\
     Hippocampus & $0.021\pm0.012$ & $0.041\pm0.026$ & $0.020\pm0.012$ \\
     Amygdala & $0.011\pm0.009$ & $0.010\pm0.001$ & $0.025\pm0.012$ \\
     Lat. Ventricles & $0.012\pm0.005$ & $0.015\pm0.010$ & $0.032\pm0.025$ \\
     White Matter & $0.005\pm0.004$ & $0.012\pm0.005$ & $0.005\pm0.004$ \\
     Grey Matter & $0.011\pm0.011$ & $0.019\pm0.006$ & $0.009\pm0.006$ \\
\hline
	\end{tabular}
  \label{tab:malc_consis}
\end{table}

\noindent
\textbf{Importance of View Aggregation: }
To evaluate the impact of view aggregation, we conducted 4 experiments with our model on 15 test volumes of MALC datasets: (i) Only coronal model, (ii) Only axial model, (iii) Aggregation of coronal and axial models, and (iv) Aggregation of coronal, axial and sagittal models. The results are reported in Table~\ref{tab:malc_viewagg}. We can observe that coronal is the best view. Aggregation with axial increases performance, with the best performance aggregating all the orthogonal views ($p<0.05$).

\begin{table}[t]
\centering
\caption{Segmentation performance for different views and with different view aggregation.}
  \begin{tabular}{|p{1.7in}|p{1.2in}|}
    \hline
     \textbf{Views} & \textbf{Mean Dice score} \\ \hline
     Only Coronal & $0.895\pm0.055$  \\
     Only Axial & $0.879\pm0.062$  \\
     Coronal + Axial & $0.897\pm0.052$  \\
     Coronal + Axial + Sagittal & $\mathbf{0.901\pm0.045}$ \\
\hline
	\end{tabular}
  \label{tab:malc_viewagg}
\end{table}

\subsection{Evaluation of segmentation accuracy with training and testing on different dataset}
\label{sec:diffdata}
In the following experiments, we evaluate the generalizability of QuickNAT by applying the network on datasets that have not been used for training. We increase the number of training scans from MALC from 15 to 28 and compare to FreeSurfer~\citep{fischl_freesurfer} and FSL~\citep{ashburner_unified_2005}, which are the most frequently used tools for neuroanatomical reconstruction. Note that manual annotations of the training/testing datasets follow the same protocol as the FreeSurfer atlas, defined by the Center for Morphometric Analysis at Massachusetts General Hospital. We assess accuracy with respect to the age of subjects, the presence of disease (Alzheimer's disease, AD), and the magnetic field strength (3.0T/1.5T). Finally, we evaluate the sensitivity of automated segmentation in comparison to manual segmentation in group analyses. 

\subsubsection{Experiment 2: ADNI-29}
\label{sec:adni29}
In this experiment, we test whole brain segmentation on 29 scans from the Alzheimer's Disease Neuroimaging Initiative (ADNI) that were manually annotated by Neomorphometrics Inc. ADNI~\citep{jack_alzheimers_2008} is one of the largest longitudinal neuroimaging studies to date.
ADNI-29 contains whole brain segmentations for 14 AD patients and 15 controls from the ADNI dataset~\citep{jack_alzheimers_2008}. The dataset contains 15 1.5T scans and 14 3.0T scans. Beside measuring segmentation accuracy, we also use this dataset to evaluate the performance of QuickNAT in group analysis by computing effect sizes and p-values. We segmented all  29 scans by applying the already trained QuickNAT model.

\begin{figure*}[t]
\centering
\includegraphics[width=0.999\textwidth]{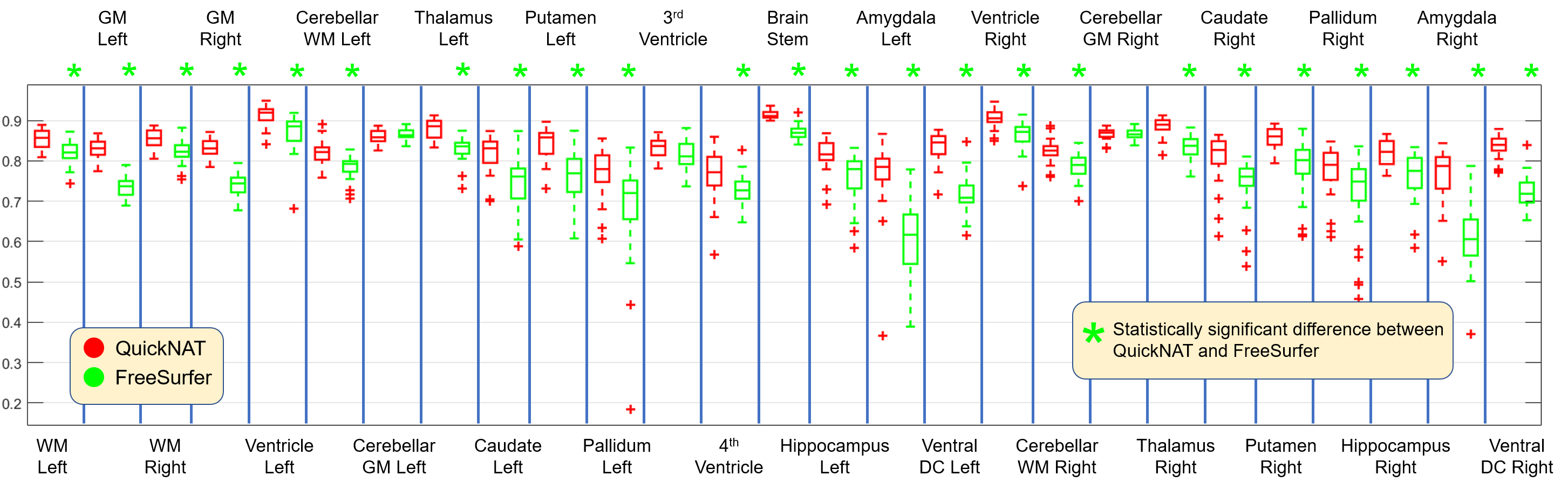}
\caption{Box-plot of Dice scores comparing QuickNAT with FreeSurfer on ADNI-29 Dataset consisting of 15 Control and 14 AD subjects for all the 27 structures. Statistical significance ($p < 0.05$) in comparison to QuickNAT is indicated by a star symbol ($\star$). The p-values were estimated using two-sided Wilcoxon rank-sum test. WM indicates White Matter and GM indicates Grey Matter. }
\label{fig:adni_box}
\end{figure*}

The mean Dice score on the dataset for FreeSurfer is $0.778\pm0.097$ and for QuickNAT $0.841\pm0.064$. QuickNAT outperforms FreeSurfer by 6\% points in Dice score with $p<10^{-7}$. Also, a structure-wise comparison is provided in Fig.~\ref{fig:adni_box}, where QuickNAT has significantly higher dice score than FreeSurfer for 24 out of 27 structures. We also deployed our `Pre-trained' network (trained with FreeSurfer auxiliary labels on IXI dataset) on this dataset, which resulted in a mean Dice score of $0.789\pm0.093$. Interestingly, this network, only trained on FreeSurfer annotations, achieves a $1.1\%$ points higher Dice score than FreeSurfer itself.

We evaluated the effectiveness of our proposed training strategy of first pre-training followed by fine-tuning in Experiment 1 (Table~\ref{tab:malc}) on MALC dataset. On ADNI-29 which is an unseen dataset, we observed a similar trend where fine-tuning lead to $6\%$ points increase in global Dice score. We conducted another experiment, where we trained QuickNAT on the union of the IXI and MALC dataset. When applied to ADNI-29 we obtain a global Dice score of $0.814\pm0.080$, which is $3\%$ points less than our proposed final framework. This is evidence that a two step pre-training from scratch and fine-tuning is better than training on the combined data.

\begin{figure}[t]
\centering
\includegraphics[width=0.48\textwidth]{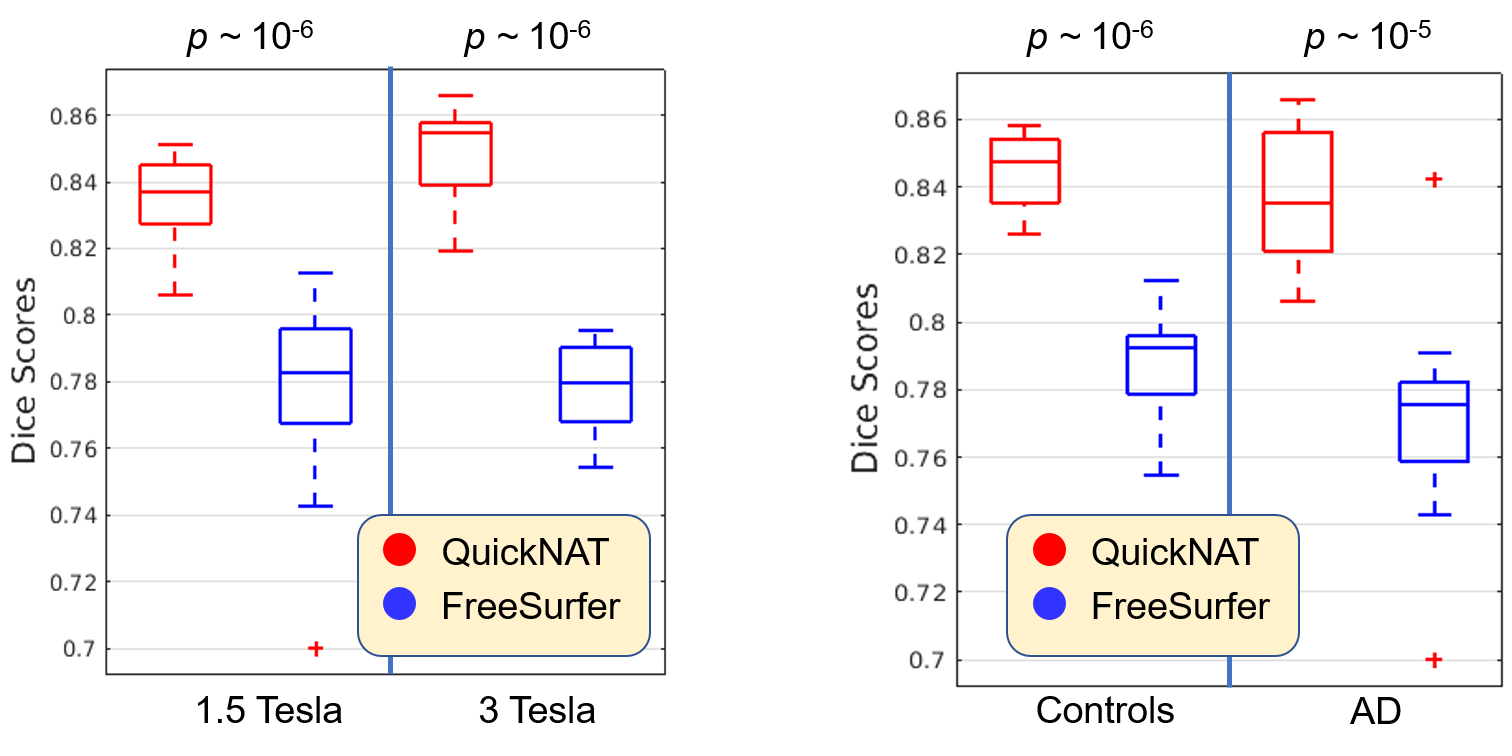}
\caption{Boxplot of Dice scores. To the left, Comparison of Dice scores for 1.5 Tesla and 3.0 Tesla field strengths for QuickNAT and FreeSurfer on ADNI-29. 15 scans were acquired at 1.5T and 14 scans at 3T. To the right, Comparison of Dice scores for AD and controls for QuickNAT and FreeSurfer on ADNI-29. 15 scans were acquired from controls and 14 from AD. The p-values were estimated using two-sided Wilcoxon rank-sum test. }
\label{fig:adni_fs}
\end{figure}

To evaluate the robustness of QuickNAT across scans acquired from scanners with different field strengths (1.5T/ 3.0T), we compared the Dice score across the two groups. We did the same with FreeSurfer. The results are shown in Fig.~\ref{fig:adni_fs}. We observe that for both groups, QuickNAT outperforms FreeSurfer. We conducted the same experiment with Control and AD patients as groups to observe robustness to pathologies. The results are presented in Fig.~\ref{fig:adni_fs}, where also we observe a superior performance of QuickNAT.

Tables~\ref{tab:adni_effect} reports the effect sizes for the group analysis by comparing AD patients with controls. We only report values for hippocampus and amygdala due to their important role in AD pathology. The analyses are performed on volume estimates, where we normalize volume estimates by the intracranial volume. For computing effect sizes, we use Hedge's g and Glass $\Delta$. Since we have less than 50 subjects for each of the groups, we use variants of these metrics, customized for small sample size. We report effect sizes and corresponding confidence intervals, computed on the manual segmentations together with those from QuickNAT and FreeSurfer. We observe that for both metrics, QuickNAT is closer to the actual estimate than FreeSurfer. 

\begin{table*}[t]
\small
\centering
\caption{Effect sizes and confidence intervals from manual segmentation, QuickNAT and FreeSurfer in terms of Hedge's g and Glass $\Delta$ for ADNI-29 dataset between controls and AD. }
  \begin{tabular}{|p{1.3in}|p{1in}|p{1in}|p{1in}|}
    \hline
      & \multicolumn{3}{c|}{Hedge's g}\\
      & \textbf{Manual} & \textbf{QuickNAT} & \textbf{FreeSurfer}\\
    \hline
    \textbf{Hippocampus Left} & $1.22 (0.37 - 1.99)$ & $1.18 (0.34 - 1.95)$ & $1.00 (0.18 - 1.76)$ \\
    \textbf{Amygdala Left} & $1.11 (0.28 - 1.88)$ & $1.16 (0.32 - 1.94)$ & $0.92 (0.12 - 1.68)$ \\ 
    \textbf{Hippocampus Right} & $1.26 (0.41 - 2.04)$ & $1.25 (0.40 - 2.03)$ & $1.06 (0.24 - 1.18)$ \\ 
    \textbf{Amygdala Right} & $1.26 (0.41 - 2.04)$ & $1.40 (0.53 - 2.20)$ & $0.96 (0.15 - 1.72)$ \\ 
    \hline
    & \multicolumn{3}{c|}{Glass $\Delta$}\\
      & \textbf{Manual} & \textbf{QuickNAT} & \textbf{FreeSurfer}\\
    \hline
    \textbf{Hippocampus Left} & $0.96 (0.11 - 1.81)$ & $0.88 (0.05 - 1.72)$ & $0.75 (0.05 - 1.57)$ \\
    \textbf{Amygdala Left} & $0.94 (0.09 - 1.79)$ & $0.93 (0.09 - 1.77)$ & $0.83 (0.00 - 1.65)$ \\ 
    \textbf{Hippocampus Right} & $1.00 (0.14 - 1.85)$ & $0.97 (0.12 - 1.82)$ & $0.83 (0.00 - 1.66)$ \\ 
    \textbf{Amygdala Right} & $1.03 (0.17 - 1.89)$ & $1.09 (0.22 - 1.95)$ & $0.79 (0.03 - 1.61)$ \\ 
    \hline
  \end{tabular}
  \label{tab:adni_effect}
\end{table*}

We also evaluate the performance of QuickNAT in finding significant associations between diagnostic groups and brain morphology. Towards this end, we use a standard linear regression model: Volume $\sim$ Age + Sex + Diagnosis, and compare the regression co-efficients of  Diagnosis, in Table~\ref{tab:adni_pvalue}. The co-efficients of QuickNAT are closer to the ones from manual segmentation. 

\begin{table}[t]
\scriptsize
\centering
\caption{Normalized volume estimates for manual, QuickNAT and FreeSurfer segmentations are used in a linear model, Volume $\sim$ Age + Sex + Disease, for the ADNI-29 dataset. The normalized regression co-efficient and p-values corresponding to variable Disease is reported below. }
  \begin{tabular}{|p{1.1in}|p{0.6in}|p{0.6in}|p{0.6in}|}
    \hline
      & \textbf{Manual} & \textbf{QuickNAT} & \textbf{FreeSurfer}\\
    \hline
    \textbf{Hippocampus Left} & $1.136 (0.0012)$ & $1.136 (0.0012)$ & $1.015 (0.0031)$ \\
    \textbf{Amygdala Left} & $1.013 (0.0058)$ & $1.124 (0.0009)$ & $0.954 (0.0068)$ \\ 
    \textbf{Hippocampus Right} & $1.157 (0.0010)$ & $1.149 (0.0011)$ & $1.061 (0.0020)$ \\ 
    \textbf{Amygdala Right} & $1.119 (0.0020)$ & $1.247 (0.0002)$ & $0.957 (0.0075)$ \\ 
    \hline
  \end{tabular}
  \label{tab:adni_pvalue}
\end{table}

\subsubsection{Experiment 3: IBSR}
\label{sec:ibsr}
\begin{figure*}[t]
\centering
\includegraphics[width=0.999\textwidth]{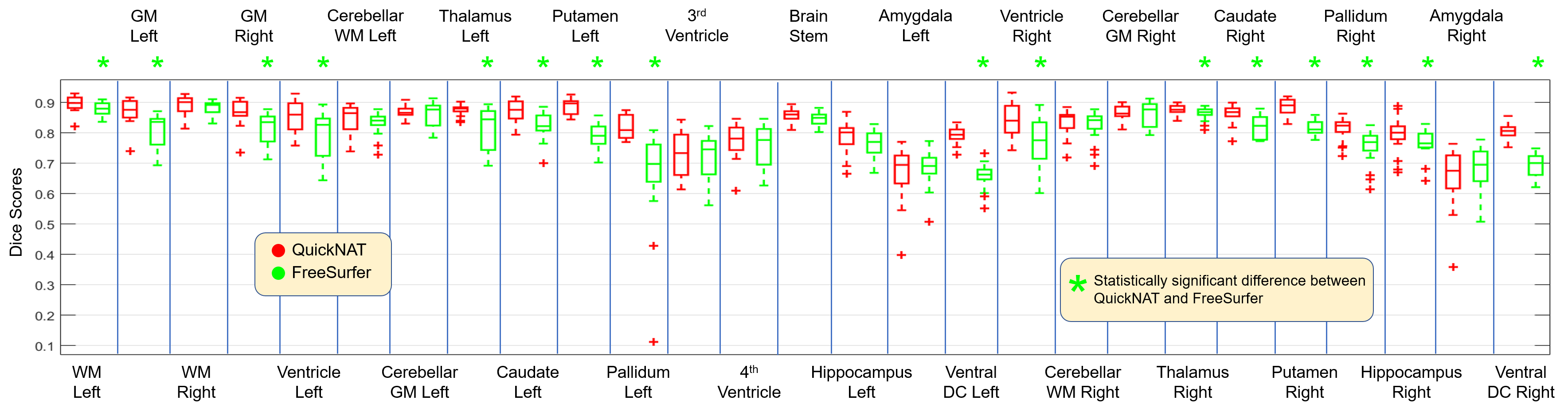}
\caption{Box-plot of Dice scores comparing QuickNAT with FreeSurfer on IBSR Dataset consisting of 18 MRI T1 scans for all the 27 structures. Statistical significance ($p < 0.05$) in comparison to QuickNAT is indicated by a star symbol ($\star$). The p-values were estimated using two-sided Wilcoxon rank-sum test. WM indicates White Matter and GM indicates Grey Matter.}
\label{fig:ibsr_box}
\end{figure*}

In this experiment, we test on 18 T1 MRI scans with manual whole brain segmentations from the Internet Brain Segmentation Repository (IBSR). Challenges of this dataset include lower resolution of 1.5mm in anterior-posterior direction (high slice thickness), low contrast, and a wide age range from 7 to 71 years. 
Results on IBSR show a higher mean Dice score for QuickNAT ($0.835\pm0.080$) than for FreeSurfer ($0.794\pm0.093$).
Fig.~\ref{fig:ibsr_box} reports structure-wise Dice scores. QuickNAT results in higher Dice scores than FreeSurfer, which is significant for the brain-wide comparison ($p<10^{-7}$ and for 16 of the 27 structures. 

Next, we compare with the CNN-based segmentation method in~\cite{dolz_3d_2017}, who segment 4 structures (thalamus, caudate, putamen, pallidum). They train their model on ABIDE (Autism Brain Imaging Data Exchange) dataset and test on IBSR, which is similar to our cross-dataset experimental setup. Table~\ref{tab:ibsr} shows that QuickNAT results in about 2-3\% points higher Dice score, except for thalamus. Note however, that~\cite{dolz_3d_2017} use the `brain.mgz' output from FreeSurfer as input (skull stripped, intensity normalized), which takes hours to generate and facilitates learning due to better standardization of the scans. In contrast, we use `orig.mgz' as input, which takes less than a second to generate. 

\begin{table}[t]
\small
\centering
\caption{Mean Dice scores of QuickNAT and~\cite{dolz_3d_2017} on IBSR for four subcortical structures segmented by~\cite{dolz_3d_2017}.}
  \begin{tabular}{|p{1in}|c|c|}
    \hline
      \textbf{Structures}& \textbf{\cite{dolz_3d_2017}} & \textbf{QuickNAT}\\
    \hline
    \textbf{Thalamus} & $0.87$ & $0.87$ \\
    \textbf{Caudate} & $0.84$ & $0.86$ \\ 
    \textbf{Putamen} & $0.85$ & $0.88$ \\ 
    \textbf{Pallidum} & $0.79$ & $0.81$ \\ 
    \hline
  \end{tabular}
  \label{tab:ibsr}
\end{table}

\subsubsection{Experiment 4: CANDI}
\label{sec:candi}
\begin{figure*}[t]
\centering
\includegraphics[width=0.999\textwidth]{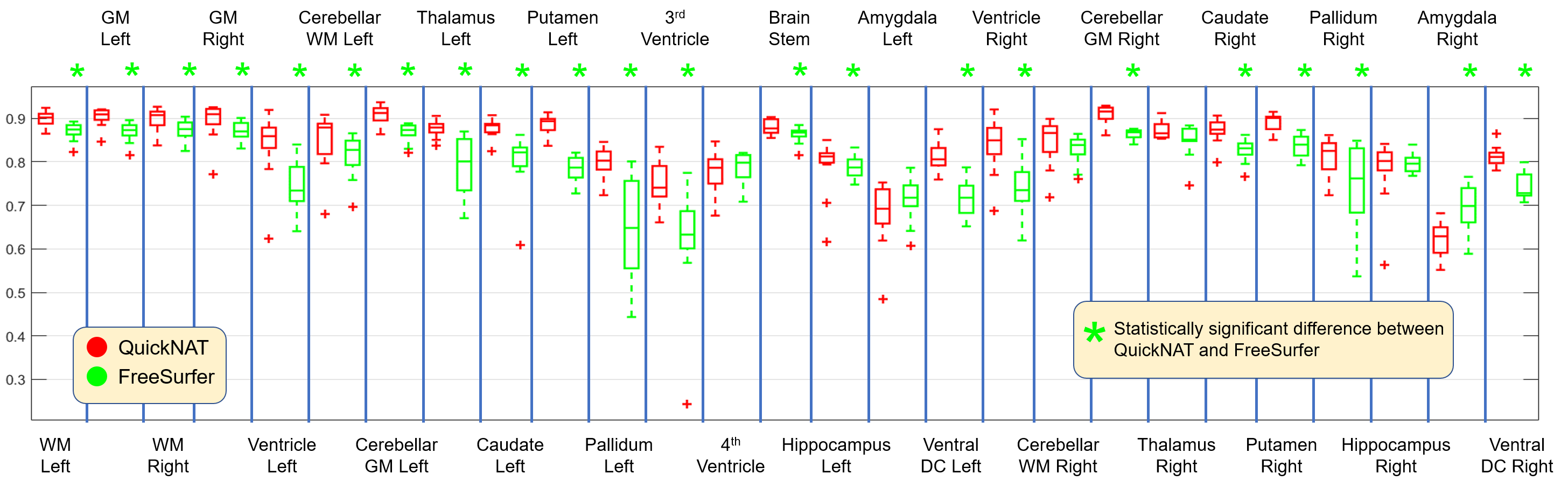}
\caption{Box-plot of Dice scores comparing QuickNAT with FreeSurfer on CANDI Dataset consisting of 13 MRI T1 scans of children for all the 27 structures. Statistical significance ($p < 0.05$) in comparison to QuickNAT is indicated by a star symbol ($\star$). The p-values were estimated using two-sided Wilcoxon rank-sum test. WM indicates White Matter and GM indicates Grey Matter. }
\label{fig:candi_box}
\end{figure*}

In this experiment, a subset of 13 MRI T1 scans from the Child and Adolescent NeuroDevelopment Initiative (CANDI) dataset~\citep{kennedy_candishare:_2012} with scans from children with psychiatric disorders were used as testing subjects. Subject's age ranges from 5 to 15 years with a mean of 10 years. Annotations were provided by Neomorphometrics Inc. We applied the trained QuickNAT on the dataset and compared against FreeSurfer. This dataset is challenging because the subjects of this particular age range are not part the training set (MALC). Here we investigate the ability of the model to generalize across previously unseen age ranges. The mean Dice scores across all structures for QuickNAT is $0.842\pm0.084$, compared to FreeSurfer with $0.798\pm0.092$, which is 5\% points lower. Fig.~\ref{fig:candi_box} reports structure-wise Dice scores. QuickNAT results in higher Dice scores than FreeSurfer, which is significant for the brain-wide comparison ($p<10^{-17}$) and for 22 of the 27 structures. 

\subsubsection{Comparison with FSL}
\label{fsl}
We also compare to FSL FIRST, which is another publicly available tool for automated segmentation of some subcortical structures for T1 MRI scans~\citep{jenkinson_fsl_2012,patenaude_bayesian_2011}. We cannot directly compare our results to FSL as it only segments $15$ structures in the brain, whereas QuickNAT segments $27$ structures. 
For a fair comparison, we selected the common 13 structures which are segmented by FSL, QuickNAT, and FreeSurfer. These are Thalamus (L+R), Caudate (L+R), Putamen (L+R), Pallidum (L+R), Hippocampus (L+R), Amygdala (L+R) and BrainStem. We reported the performance of QuickNAT, FreeSurfer and FSL on the 3 datasets with whole-brain manual annotations (ADNI-29, CANDI, IBSR). 

We observed that FSL is prone to registration errors which leads to process termination or faulty segmentations. We note that FSL was used with its default settings for all the experiments, as recommended in its instruction manual. FreeSurfer and QuickNAT did not fail during segmentation. Thus, we report results once on all scans and once on those scans that FSL did not fail in Table~\ref{tab:fsl}. The failure rate of FSL is higher on IBSR (44\%) than on ADNI (17\%) or CANDI (23\%), indicating that it is susceptible to low quality scans. Considering all scans and a sub-set of 13 structures QuickNAT outperforms both FreeSurfer and FSL by a statistically significant margin ($p<0.001$). Excluding the scans where FSL failed, QuickNAT demonstrates superior performance for ADNI-29, while FSL is best for CANDI. In IBSR, FSL outperforms QuickNAT by a small margin of 0.08\% points, which is not statistically significant. A possible reason for the good performance of FSL on CANDI may be that FSL includes scans with age range (5-15 years) in its atlas. Scans of this age range were not included in training QuickNAT. 
The performance of QuickNAT can potentially be improved on young subjects with more training data from such an age range.

\begin{table*}[t]
\small
\centering
\caption{Comparison of QuickNAT, FSL and FreeSurfer for segmentation of 13 structures common to all on ADNI-29, CANDI and IBSR Dataset. The results including and excluding scans where FSL failed are presented separately, along with the failure rate of FSL.}
  \begin{tabular}{|p{1in}|p{1in}|c|c|c|c|}
    \hline
    & \textbf{Method} & \textbf{ADNI-29} & \textbf{CANDI} & \textbf{IBSR}\\
    \hline
     \multirow{4}{*}{All scans}
     & \textbf{QuickNAT} & $\mathbf{0.825\pm0.027}$ & $\mathbf{0.819\pm0.028}$ & $\mathbf{0.820\pm0.035}$ \\
     & \textbf{FreeSurfer} & $0.745\pm0.042$ & $0.780\pm0.025$ & $0.776\pm0.025$ \\
     & \textbf{FSL} & $0.643\pm0.290$ & $0.647\pm0.369$ & $0.461\pm0.419$ \\ \cline{2-5}
     & \textbf{Failure of FSL} & 5 out of 29 (17\%) & 3 out of 13 (23\%) & 8 out of 18 (44\%) \\
     \hline
     \multirow{2}{*}{Scans where FSL}
     & \textbf{QuickNAT} & $\mathbf{0.823\pm0.027}$ & $0.817\pm0.032$ & $0.817\pm0.035$ \\
     & \textbf{FreeSurfer} & $0.745\pm0.045$ & $0.775\pm0.027$ & $0.772\pm0.022$\\
     succeeded & \textbf{FSL} & $0.775\pm0.024$ & $\mathbf{0.841\pm0.013}$ & $\mathbf{0.825\pm0.013}$ \\
    \hline
  \end{tabular}
  \label{tab:fsl}
\end{table*}

\subsubsection{Compilation of worst segmentation performance scans (Exp: 2,3,4)}
\begin{figure*}[t]
\centering
\includegraphics[width=0.999\textwidth]{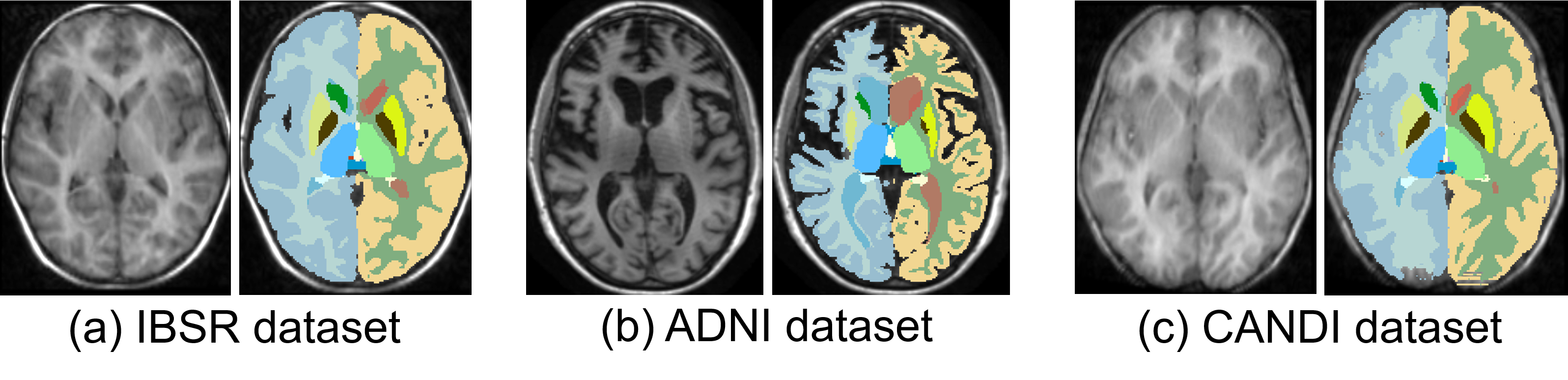}
\caption{Illustration of scans for IBSR, ADNI and CANDI with the \emph{worst segmentation} performance. The MRI scans along with QuickNAT segmentations are shown. (a) IBSR scan has motion and ringing artifacts with poor contrast, (b) ADNI scan is from a 95-year-old subject with severe AD (prominent cortical atrophy and enlarged ventricles), (c) CANDI scan is from a 5-year-old subject, with severe motion artifacts and very low contrast.  }
\label{fig:worstcase}
\end{figure*}

In this section, we visualize the subjects with lowest segmentation accuracy (mean Dice score) for all datasets with whole-brain annotations and cross-dataset evaluation (IBSR, ADNI-29, CANDI), to identify the limits of QuickNAT. Segmentations are shown in Fig.~\ref{fig:worstcase}(a-c). 
The scan with the worst performance in IBSR dataset has a mean Dice score of $0.78$, which is 5\% points less than the overall Dice of the dataset. We observe that the scan has motion and ringing artifacts and low contrast (Fig.~\ref{fig:worstcase}(a)), which might have impaired performance.
The worst performing ADNI-29 scan has a Dice score of $0.81$, which is 3\% points less than the overall Dice of dataset. The scan is from a 95-year-old patient with severe AD. The scan shows strong ringing artifacts, pronounced atrophy, and enlarged ventricles, shown in Fig.~\ref{fig:worstcase}(b). Such pathological data were not used for training, still QuickNAT generalizes well to such cases.
The worst performing CANDI scan has a Dice score of $0.77$, which is 7\% points lower than the Dice score for the overall dataset. This is from a patient aged 5. The scan has severe motion artifacts and low contrast as shown in Fig.~\ref{fig:worstcase}(c). Looking at the segmentation results of QuickNAT on all of these challenging cases, we observe small errors, but it is striking that overall decent segmentation performance is decent.

\subsubsection{Experiment 5: HarP Dataset}
\begin{figure}[t]
\centering
\includegraphics[width=0.48\textwidth]{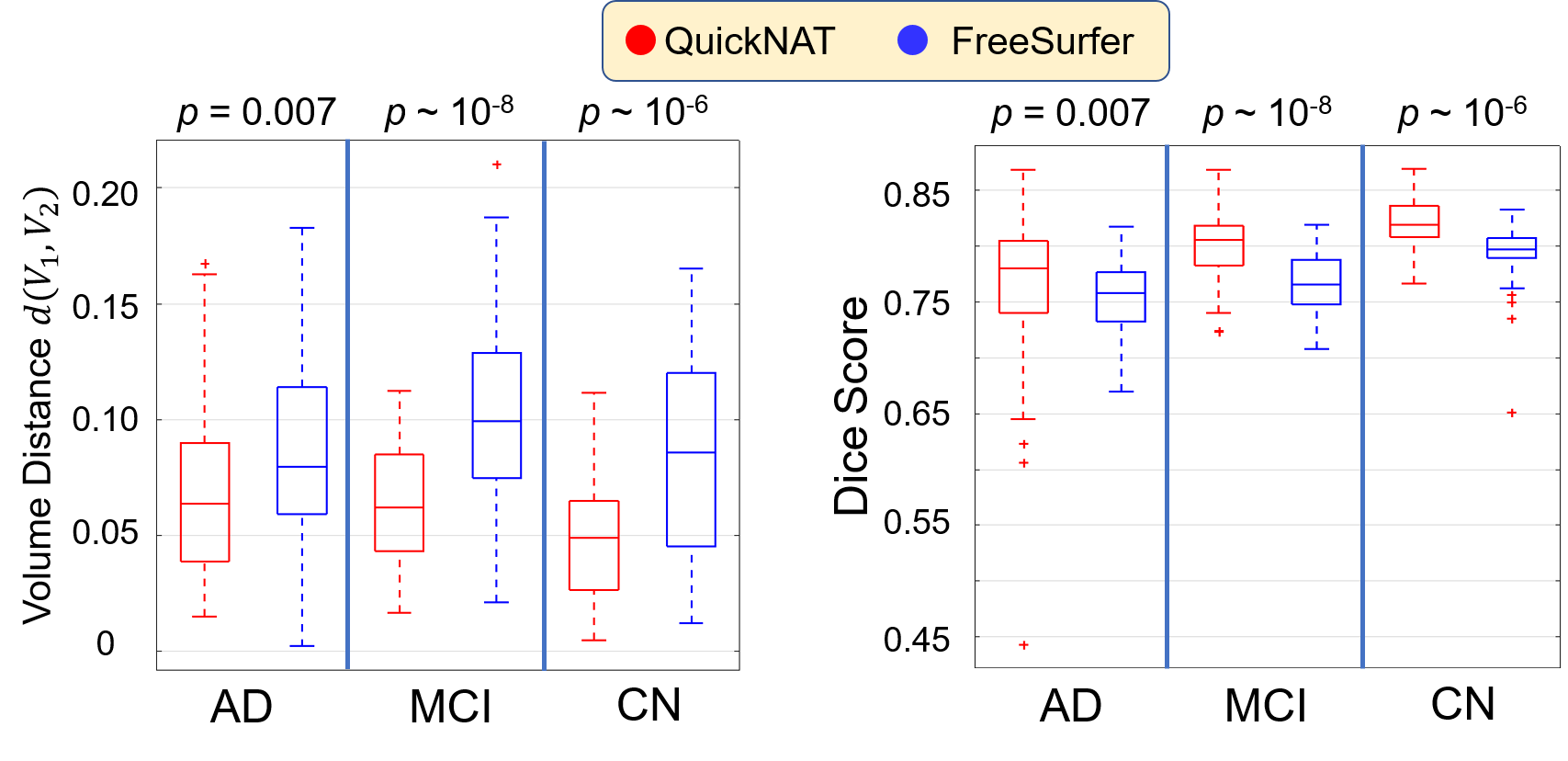}
\caption{The figure shows hippocampus segmentation performance on the HarP dataset, in terms of Dice score and volume distance for different diagnoses (42 Controls, 44 MCI, and 45 AD). Higher Dice score and correlation indicates better performance, while lower volume distance indicates better performance. The p-values were computed using two-sided Wilcoxon rank-sum test.}
\label{fig:harp}
\end{figure}

We evaluate the hippocampus segmentation in the presence of dementia on a large dataset with manual annotations following the Harmonized Protocol (HarP) for hippocampus segmentation \citep{boccardi_training_2015}, developed by the European Alzheimer's Disease Consortium and ADNI. 
Similar to ADNI-29, it is a subset of the ADNI dataset~\citep{jack_alzheimers_2008}. 
Left and right hippocampi were segmented for $131$ subjects, balanced for controls, mild cognitive impairment (MCI), and AD.
The hippocampus is an important brain structure, whose volume and shape changes are important biomarkers for disease and aging~\citep{bartsch_hippocampus_2015}. 
Segmenting the hippocampus is challenging because of small or absent signal gradient between it and its adjacent regions.
The HarP dataset contains $131$ MRI T1 scans, which are balanced for controls, MCI and AD (42 CN, 44 MCI, 45 AD). The challenges associated with this dataset are: (i) subjects with neurodegenerative disease, and (ii) variations in the manual annotation protocol. 
Across all diagnostic groups (CN, MCI, AD), the volume distance to the manual segmentation is significantly lower for QuickNAT than FreeSurfer and the Dice scores are significantly higher (Fig.~\ref{fig:harp}). 

\subsection{Evaluation of segmentation reliability}
\label{sec:segreg}
In the following experiments, we evaluate the segmentation reliability of QuickNAT with different experiments detailed below.

\subsubsection{Experiment 6: ALVIN Dataset}
\begin{figure}[t]
\centering
\includegraphics[width=0.45\textwidth]{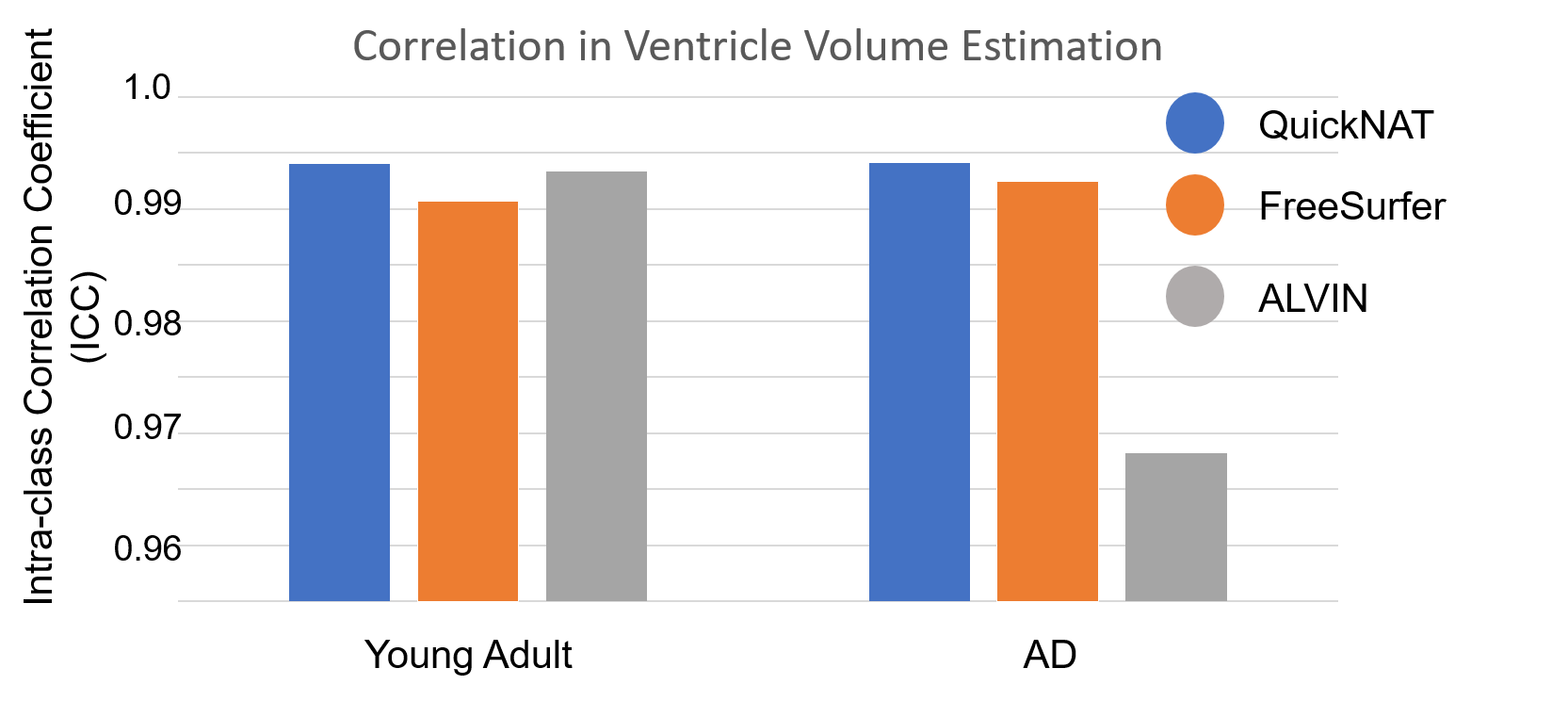}
\caption{The figure illustrates the performance of QuickNAT for ventricle volume estimation after segmentation and compares to FreeSurfer and ALVIN, on the dataset introduced by ALVIN. The performance is evaluated by estimating intra-class correlation (metric used in \cite{kempton_comprehensive_2011}) for both young adults (7 scans) and AD subjects (9 scans).}
\label{fig:alvin}
\end{figure}

In this experiment, we follow a comprehensive testing protocol for MRI neuroanatomical segmentation techniques as proposed for lateral ventricle segmentation, termed ALVIN~\citep{kempton_comprehensive_2011}. The dataset consists of 7 young adults and 9 patients with Alzheimer's disease. Lateral ventricles have been manually annotated at two time points to observe intra-observer variability and their volumes were reported. We compute volumes from QuickNAT segmentations and follow the evaluation protocol by reporting intra-class correlation coefficient (ICC)~\citep{kempton_comprehensive_2011}, in addition to volume distance (Methods). We compare with results of FreeSurfer and ALVIN reported in~\cite{kempton_comprehensive_2011}. On control subjects and AD patients, QuickNAT shows the best performance (Fig.~\ref{fig:alvin}). 

\subsubsection{Experiment 7: TRT Dataset}
The dataset was released to test reliability of automated whole-brain segmentation algorithms in estimating volumes for some important brain structures using test re-test~\citep{maclaren_reliability_2014}. They acquired 120 MRI T1 scans from 3 subjects (40 scans per subject). The scans were acquired over 20 sessions (2 scans per session). All the scans were acquired within a period of 30 days. \cite{maclaren_reliability_2014} analyzed the coefficient of variation in volume estimates within one session (intra session $CV_s$) and the total variance over all 40 scans ($CV_t$). 
\revision{The metric co-efficient of variance, provides an extent of variability to the mean value.} The total coefficient of variation is computed as $CV_t = \frac{{\sigma}_t}{{\mu}_t}$, where $\sigma_t$ and $\mu_t$ are the standard deviation and mean of the estimates. This global variance considering all the estimates across session provides inter-session $CV_t$. The intra-session co-efficient of variation $CV_s$ is estimated by the root mean square of all the co-efficient of variance per session.
Ideally as atrophy is almost negligible within the period of 30 days, the coefficient of variation in estimates should be zero. The lower the estimate, the better the estimator. We compare QuickNAT with FreeSurfer in this regard. The 8 structures considered in this experiment are consistent with the ones reported in~\cite{maclaren_reliability_2014}. The results of the experiment are reported in Table~\ref{tab:retest}. Both coefficients of variation $CV_s$ and $CV_t$ are within a tolerable range of less than 2\% for QuickNAT, but variations are quite high for FreeSurfer for some structures like thalamus (6\%), pallidum (5\%),  amygdala (5\%) and putamen (4\%). In volume estimation of cerebral WM, FreeSurfer exhibits a better performance than QuickNAT with ($CV<1\%$). A possible reason might be the sophisticated surface processing and correction stage in FreeSurfer that follows the initial segmentation~\citep{dale_cortical_1999,fischl_cortical_1999}.

Finally, having such low $CV$ estimates makes QuickNAT a promising tool for group analysis studies over large datasets with reliable estimation of biomarkers. It can also be effectively used in processing longitudinal scans to model disease progression. We excluded comparison with FSL in this experiment due to FSL registration errors in some volumes. 

\begin{table*}[t]
\small
\centering
\caption{Variation in volume measurement per structure. QuickNAT is compared against FreeSurfer in terms of intra-session coefficient of variation ($CV_s$), inter-session total coefficient of variation ($CV_t$) and the absolute difference between them. Also, the mean volume estimates per structure are reported. It must be noted that volumes of both left and right hemispheres are combined to estimate the total volume for each structure. The estimates from FreeSurfer were taken from \cite{maclaren_reliability_2014}.}
  \begin{tabular}{|p{1.3in}|c|c|c|c|c|c|c|c|}
    \hline
      Structures & \multicolumn{2}{c|}{Mean Vol. (ml)} & \multicolumn{2}{c|}{Intra-session $CV_s$ (\%)} & \multicolumn{2}{c|}{Inter-session $CV_t$ (\%)} & \multicolumn{2}{c|}{$|CV_s-CV_t|$ (\%)}\\ \cline{2-9}
      & \textbf{FS} & \textbf{Q-NAT} & \textbf{FS} & \textbf{Q-NAT} & \textbf{FS} & \textbf{Q-NAT} & \textbf{FS} & \textbf{Q-NAT}\\
    \hline
    Hippocampus &	8.90 &	7.56 &	2.77 &	\textbf{0.73} &	2.92 &	\textbf{0.80} &	0.16 &	\textbf{0.09} \\
Lateral Ventricles &	10.10 &	14.42 &	\textbf{1.58} &	2.33 &	3.40 &	\textbf{3.04} &	1.82 &	\textbf{0.95}\\
Amygdala &	3.80 &	2.18 &	4.69 &	\textbf{1.91} &	5.21 &	\textbf{2.39} &	0.53 &	\textbf{0.50}\\
Putamen &	11.60 &	8.89 &	4.04 &	\textbf{0.71} &	3.92 &	\textbf{0.85} &	\textbf{0.13} &	0.17\\
Pallidum &	3.20 &	3.36 &	5.25 &	\textbf{1.32} &	5.42 &	\textbf{1.42} &	0.17 &	\textbf{0.12}\\
Caudate &	7.40 &	6.82 &	1.54 &	\textbf{1.02} &	1.58 &	\textbf{1.14} &	\textbf{0.04} &	0.18\\
Thalamus &	12.90 &	16.22 &	5.98 &	\textbf{0.77} &	6.06 &	\textbf{0.93} &	\textbf{0.08} &	0.19\\
Cerebral WM &	496.60 &	403.70 &	\textbf{0.88} &	1.98 &	\textbf{0.87} &	1.91 &	\textbf{0.00} &	0.07\\
    \hline
  \end{tabular}
  \label{tab:retest}
\end{table*}

\subsubsection{Experiment 8: HTP Dataset}
In this experiment, we evaluated the reliability and robustness in estimating volumes across scans acquired from multiple centers, from the Human Travelling Phantom (HTP) dataset~\citep{magnotta_multicenter_2012}.
The dataset includes scans from 5 healthy subjects travelling to 8 different medical centers in the USA. All scans were acquired within a period of 30 days, such that atrophy of structures due to normal aging is negligible. Each of the 8 imaging centers used MRI scanners manufactured by different vendors, different gradient specifications, different number of channels in the head coil etc. Ideally, the coefficient of variation ($CV$) of volume estimates across the sites should be zero; the lower the estimate, the more reliable and robust is the segmentation algorithm. A detailed explanation of the experimental setup is provided in~\cite{magnotta_multicenter_2012}. Overall, the dataset is challenging for segmentation, because it is very heterogeneous with strong variation of data quality across sites, where scans from some sites exhibit strong motion artifacts. 
We compared QuickNAT with FreeSurfer and reported the results in Table~\ref{tab:htp}. QuickNAT showed more robustness for hippocampus, putamen, pallidum, and thalamus, while FreeSurfer is better in lateral ventricles, amygdala, caudate, and cerebral WM. Overall, this challenging experiment demonstrated that QuickNAT and FreeSurfer are equally robust. 

\begin{table*}[t]
\small
\centering
\caption{Coefficient of variation ($CV$) in volume estimation for the 8 structures for each subject, using QuickNAT (QN) and FreeSurfer (FS). Also, RMS $CV$ per structures for all the subjects is presented.}
  \begin{tabular}{|p{1.3in}|c|c|c|c|c|c|c|c|c|c|c|c|}
    \hline
    & \multicolumn{10}{c|}{Subject ID with $CV$} & \multicolumn{2}{c|}{RMS}\\ \cline{2-11}
    Structures &  \multicolumn{2}{c|}{1} & \multicolumn{2}{c|}{2} & \multicolumn{2}{c|}{3} & \multicolumn{2}{c|}{4} & \multicolumn{2}{c|}{5} & \multicolumn{2}{c|}{$CV$}\\ \cline{2-13}
    & \textbf{FS} & \textbf{QN} & \textbf{FS} & \textbf{QN} & \textbf{FS} & \textbf{QN} & \textbf{FS} & \textbf{QN} & \textbf{FS} & \textbf{QN} & \textbf{FS} & \textbf{QN} \\ \hline
    Hippocampus	&	8.49	&	4.66	&	1.41	&	1.09	&	2.39	&	2.18	&	3.57	&	1.19	&	2.69	&	4.06	&	4.47	&	\textbf{3.02} \\
Lateral Ventricles	&	8.28	&	14.8	&	5.25	&	8.43	&	2.37	&	6.32	&	2.85	&	6.62	&	12.9	&	15.1	&	\textbf{7.45}	&	10.9 \\
Amygdala	&	4.1	&	10.4	&	2.25	&	5.08	&	5.33	&	4.98	&	2.16	&	2.99	&	5.58	&	5.31	&	\textbf{4.15}	&	6.29 \\
Putamen	&	9.36	&	5.76	&	5.66	&	1.89	&	5.83	&	1.3	&	5.18	&	1.8	&	4.29	&	5.08	&	6.31	&	\textbf{3.63} \\
Pallidum	&	9.34	&	5.26	&	7.59	&	1.49	&	9.05	&	1.59	&	7.99	&	1.22	&	5.27	&	3.85	&	7.98	&	\textbf{3.12} \\
Caudate	&	5.12	&	15.3	&	2.83	&	5.72	&	3.13	&	1.47	&	2.45	&	4.27	&	3.01	&	6.75	&	\textbf{3.44}	&	8.18 \\
Thalamus	&	2.74	&	1.43	&	4.21	&	1.11	&	2.7	&	2.55	&	2.42	&	2.36	&	5.33	&	3.49	&	3.65	&	\textbf{2.35} \\
Cerebral WM	&	3.09	&	3.43	&	1.64	&	3.45	&	2.59	&	4.8	&	2.2	&	3.79	&	4.14	&	4.28	&	\textbf{2.86}	&	3.99 \\

    \hline
  \end{tabular}
  \label{tab:htp}
\end{table*}  

\section{Discussion}    
\subsection{Comparison with Deep Learning Approaches}
Recently, convolutional neural networks have been proposed for brain segmentation~\citep{chen_voxresnet:_2017, dolz_3d_2017, fedorov_almost_2017, wachinger_deepnat:_2017, moeskops2016automatic}.
DeepNAT~\citep{wachinger_deepnat:_2017} reported competitive results on the MALC data, but as shown in Table~\ref{tab:malc}, QuickNAT yields significantly higher accuracy, while requiring only seconds (Fig.~\ref{fig:speed}).
\cite{dolz_3d_2017} proposed a network for segmenting 8 structures based on skull-stripped and intensity normalized images, which facilitates the learning process but requires hours for processing. Nevertheless, a comparison on IBSR demonstrated higher accuracy for QuickNAT (Table~\ref{tab:ibsr}). The comparison to VoxResNet~\citep{chen_voxresnet:_2017} is not directly possible, because only 3 structures were segmented. MeshNET~\citep{fedorov_almost_2017} only uses FreeSurfer segmentations for training and testing, without any manual annotations, which makes it complicated to assess the actual performance. The methods in \cite{dolz_3d_2017} and \cite{fedorov_almost_2017} have limited comparison to existing segmentation approaches, while we compare with FreeSurfer, FSL, PICSL, SpatialSTAPLE, U-Net, FCN, and DeepNAT using an identical experimental setup. Notably, none of the methods have been used in a cross-dataset evaluation, i.e., training and testing on separate datasets, except for \cite{dolz_3d_2017}. By evaluating QuickNAT on 8 different datasets and performing reliability study on 3 datasets, we have presented the most comprehensive evaluation of a convolutional neural network for brain segmentation so far. 

In addition to the above mentioned articles, CNNs have also been proposed for the segmentation of pathological structures like brain tumors~\citep{pereira2016brain, havaei2017brain, kamnitsas2017efficient} or lesions~\citep{brosch2016deep, ghafoorian2017location, valverde2017improving}.

\subsection{Pre-training with Auxiliary Labels}
Although deep learning models have been shown to be highly effective, they are highly complex and require large annotated data for effective training~\citep{lecun_deep_2015}. Access to abundant annotated training data is challenging for medical applications due to the high cost of creating expert annotations. The problem is more prominent for F-CNNs, where each slice corresponds to one data point, in contrast to patch based approaches, where millions of patches can be extracted from a volume~\citep{wachinger_deepnat:_2017}. To address this issue, we introduced a training strategy that leverages large unlabeled data and small manual data to effectively train our fully convolutional model. We used FreeSurfer to automatically create segmentations from unlabeled data, which serve as auxiliary labels to pre-train our model. This pre-trained model, which mimics FreeSurfer, is fine-tuned with small manually annotated data to get the final model. Our results have shown that a model trained with this new strategy significantly outperforms a model that is only trained on manual data. Although we have demonstrated the application to brain segmentation, the proposed training strategy is generic and can be effectively used for other segmentation tasks as well. In a parallel research work, FreeSurfer generated labels were used to train a model for multiple cohorts for hippocampus segmentation, showing promising results~\citep{thyreau2018segmentation}.

In Sec.~\ref{sec:adni29}, we observed another very interesting aspect of pre-training. On the ADNI-29 dataset, the pre-trained network achieved a higher accuracy than FreeSurfer itself. This is quite surprising given the fact that pre-training was conducted with annotations generated from FreeSurfer only. 
In other words, it seems that the network imitating FreeSurfer can perform better than FreeSurfer itself. 
 The reason for such a behavior could be the large amount of data (IXI Dataset) that the pre-trained model has seen and that it learned to generalize from the noisy auxiliary annotations, emphasizing the potential of pre-training. 

\subsection{Architecture Design}
The architecture of QuickNAT has been tailored to address the challenges associated with whole brain segmentation. The fully convolutional architecture offers faster processing and larger context than patch-based DeepNAT~\citep{wachinger_deepnat:_2017}, because all the voxels in a slice are labelled simultaneously. The dense connections within every encoder and decoder block promote feature re-usability in the network~\citep{densenet}, which optimizes model complexity by avoiding learning of similar feature maps in different layers. In the decoder blocks, upsampling is done using unpooling layers instead of convolutions, which does not involve any learnable parameters and enforces spatial consistency; an essential aspect for segmenting small subcortical structures. The network is learned by optimizing a combined weighted logistic and Dice loss function with stochastic gradient descent. To tackle class imbalance and provide reliable contour estimation of the structures, we up-weighted under-represented classes using median frequency balancing and emphasized anatomical boundaries. Our results have shown the significant improvement of the QuickNAT architecture, compared to state-of-the-art U-Net and FCN models.

\subsection{Segmentation Accuracy}
We have demonstrated the high accuracy of QuickNAT on a comprehensive set of 5 experiments that cover a wide range of variations in acquisition parameters and neuroanatomy. In experiments on the MALC dataset, we demonstrated that QuickNAT provides segmentations with similar accuracy and inter-run consistency to the best atlas-based methods (Table.~\ref{tab:malc} and Table.~\ref{tab:malc_consis}). In experiments with ADNI-29, we demonstrated the robustness of QuickNAT with respect to pathology and magnetic field strengths (Fig.~\ref{fig:adni_fs}). Moreover, effect sizes from QuickNAT are more similar to those from manual segmentations than FreeSurfer (Table~\ref{tab:adni_effect}). In experiments with IBSR, we demonstrated the robustness to data with lower resolution and low contrast (Fig.~\ref{fig:ibsr_box}). High segmentation accuracy on scans from young subjects was demonstrated on the CANDI dataset (Fig.~\ref{fig:candi_box}) and for hippocampus segmentation on the HarP dataset (Fig.~\ref{fig:harp}). Finally, we compared to FSL FIRST on a subset of structures that are identified by both (Table.~\ref{tab:fsl}). Notably, QuickNAT has not failed on any scan across all datasets and has not produced a single segmentation that had to be rejected.  

\subsection{Segmentation Reliability}
In an additional set of 3 experiments, we evaluated the reliability of QuickNAT. We measured high reliability in lateral ventricle segmentation by following the testing protocol ALVIN. We observed high consistency for the segmentation of 8 structures on test-retest data with less than 2\% variation on most brain structures (Table.~\ref{tab:retest}). The test-retest data was acquired on the same scanner. We extended the evaluation to a more challenging dataset, where the same subject was scanned in various machines at different sites. As expected, the variation increased in this setup, but the reliability was comparable to FreeSurfer. The high reliability of QuickNAT compared to FreeSurfer is surprising, because the atlas used in FreeSurfer provides a strong spatial prior, which tends to improve reliability. However, our results showed that the more unconstrained, deep learning based segmentation can achieve higher reliability.


\subsection{Limitations}
As any brain segmentation method, QuickNAT has limitations. On the data of the human traveling phantom (HTP Dataset), we observed an increase in variance across centers. In a detailed investigation, we found that scans from one of the sites (Dartmouth) had strong motion artifacts, which deteriorated our segmentation performance, and in turn increased the variance. Motion artifacts present a challenge to many image processing tasks. If more than one source volume exists, motion correction could be applied, as is done in the FreeSurfer pipeline. We have experimented with scans from subjects in the age range 5 to 95. Outside this age range, additional experiments need to be conducted. Furthermore, we have shown scans with the worst segmentation performance in Fig.~\ref{fig:worstcase}, illustrating the limits of QuickNAT. Another limitation of QuickNAT is that it cannot deal with tissue classes that are not part of the training set, e.g., tumors. For QuickNAT to also work on tumor cases, we would need training data, where the tumor is annotated together with all the brain structures. To the best of our knowledge, such a dataset is not publicly available.

\section{Conclusion}
We have introduced QuickNAT, a deep learning based method for brain segmentation that runs in seconds, achieving superior performance with respect to existing methods and being orders of magnitudes faster in comparison to patch-based CNNs and atlas-based approaches.
We have demonstrated that QuickNAT generalizes well to other, unseen datasets (training data different to testing) and yields high segmentation accuracy across diagnostic groups, scanner field strengths, and age, while producing highly consistent results. 
This high segmentation accuracy enhances group analyses by enabling effect sizes and significance values that better match those of manual segmentations. 
Also, with high test-retest accuracy it can be effectively used for longitudinal studies. QuickNAT can be highly impactful because of its fast processing time and robustness to neuroanatomical variability, allowing for an almost instantaneous access to accurate imaging biomarkers.

\section*{Acknowledgment}
Support for this research was provided in part by the Bavarian State Ministry of Education, Science and the Arts in the framework of the Centre Digitisation.Bavaria (ZD.B). We thank Neuromorphometrics Inc. for providing manual annotations, neuroimaging initiatives for sharing data, and NVIDIA corporation for GPU donation.
We would also like to thank Dr. Sebastian P\"{o}lsterl for proofreading the manuscript and providing feedback.
Data collection and sharing was funded by the Alzheimer's Disease Neuroimaging Initiative (ADNI) (National Institutes of Health Grant U01 AG024904) and DOD ADNI (Department of Defense award number W81XWH-12-2-0012). ADNI is funded by the National Institute on Aging, the National Institute of Biomedical Imaging and Bioengineering, and through generous contributions from the following:
Alzheimer's Association; Alzheimer's Drug Discovery Foundation; Araclon Biotech; BioClinica Inc.; Biogen Idec Inc.; Bristol-Myers Squibb Company; Eisai Inc.; Elan Pharmaceuticals, Inc.; Eli Lilly and Company; EuroImmun; F. Hoffmann-La Roche Ltd and its affiliated company Genentech, Inc.; Fujirebio; GE Healthcare; IXICO Ltd.; Janssen Alzheimer Immunotherapy Research \& Development, LLC ; Johnson \& Johnson Pharmaceutical Research \& Development LLC; Medpace, Inc; Merck \& Co., Inc.; Meso Scale Diagnostics, LLC; NeuroRx Research; Neurotrack Technologies; Novartis Pharmaceuticals Corporation; Pfizer Inc.; Piramal Imaging; Servier; Synarc Inc.; and Takeda Pharmaceutical Company. 
The Canadian Institutes of Health Research is providing funds to support ADNI clinical sites in Canada. Private sector contributions are facilitated by the Foundation for the National Institutes of Health (www.fnih.org). The grantee organization is the Northern California Institute for Research and Education, and the study is coordinated by the Alzheimer's Disease Cooperative Study at the University of California, San Diego. ADNI data are disseminated by the Laboratory for Neuro Imaging at the University of Southern California.

\section*{Appendix}
\textbf{List of classes:}
The brain structures segmented by QuickNAT are: (1) Cortical White Matter Left, (2) Cortical Grey Matter Left, (3) Cortical White Matter Right (4) Cortical Grey Matter Right, (5) Lateral Ventricle Left, (6) Cerebellar White Matter Left, (7) Cerebellar Grey Matter Left, (8) Thalamus Left, (9) Caudate Left, (10) Putamen Left, (11) Pallidum Left, (12) $3^{rd}$ ventricle, (13) $4^{th}$ ventricle, (14) Brainstem, (15) Hippocampus Left, (16) Amygdala Left, (17) Ventral DC Left, (18) Lateral Ventricle Right, (19) Cerebellar White Matter Right, (20) Cerebellar Grey Matter Right, (21) Thalamus Right, (22) Caudate Right, (23) Putamen Right, (24) Pallidum Right, (25) Hippocampus Right, (26) Amygdala Right, (27) Ventral DC Right.

\textbf{Label remapping strategy:}
QuickNAT segments 27 brain structures with IDs 1 to 27 as indicated above. For training, testing and evaluation purposes, we map the FreeSurfer labels and Manual Labels (provided by Neuromorphometrics Inc.) consistent to that of the QuickNAT IDs. The ID mapping strategy is detailed in Tab.~\ref{tab:label_remap}. For FreeSurfer, the mapping IDs are corresponding to `\texttt{aseg.mgz}', which does not contain cortical parcellations. For manual annotations, which has cortical parcellations, we first map all the parcels to a single cortex class. All the IDs greater than $100$ with even values are mapped to ID $210$ (Right hemisphere cortex). Similarly, all the IDs greater than $100$ with odd values are mapped to ID $211$ (left hemisphere cortex). After this the mapping to QuickNAT IDs are performed as per Tab.~\ref{tab:label_remap}.

\begin{table*}[t]
\small
\centering
\caption{Label Remapping Strategy}
  \begin{tabular}{|p{2in}|c|c|c|}
    \hline
      \textbf{Structures}& \textbf{QuickNAT} & \textbf{FreeSurfer} & \textbf{Manual}\\
    \hline
    \textbf{Cortical White Matter Left} & $1$ & $2$ & $45$ \\
    \textbf{Cortical Grey Matter Left} & $2$ & $3$ & $211$ \\
    \textbf{Cortical White Matter Right} & $3$ & $41$ & $44$ \\
    \textbf{Cortical Grey Matter Right} & $4$ & $42$ & $210$ \\
    \textbf{Lateral Ventricle Left} & $5$ & $4$ & $52$ \\
    \textbf{Cerebellar White Matter Left} & $6$ & $7$ & $41$ \\
    \textbf{Cerebellar Grey Matter Left} & $7$ & $8$ & $39$ \\
    \textbf{Thalamus Left} & $8$ & $10$ & $60$ \\
    \textbf{Caudate Left} & $9$ & $11$ & $37$ \\
    \textbf{Putamen Left} & $10$ & $12$ & $58$ \\
    \textbf{Pallidum Left} & $11$ & $13$ & $56$ \\
    \textbf{$3^{rd}$ ventricle} & $12$ & $14$ & $4$ \\
    \textbf{$4^{th}$ ventricle} & $13$ & $15$ & $11$ \\
    \textbf{Brainstem} & $14$ & $16$ & $35$ \\
    \textbf{Hippocampus Left} & $15$ & $17$ & $48$ \\
    \textbf{Amygdala Left} & $16$ & $18$ & $32$ \\
    \textbf{Ventral DC Left} & $17$ & $28$ & $62$ \\
    \textbf{Lateral Ventricle Right} & $18$ & $43$ & $51$ \\
    \textbf{Cerebellar White Matter Right} & $19$ & $46$ & $40$ \\
    \textbf{Cerebellar Grey Matter Right} & $20$ & $47$ & $38$ \\
    \textbf{Thalamus Right} & $21$ & $49$ & $59$ \\
    \textbf{Caudate Right} & $22$ & $50$ & $36$ \\
    \textbf{Putamen Right} & $23$ & $51$ & $57$ \\
    \textbf{Pallidum Right} & $24$ & $52$ & $55$ \\
    \textbf{Hippocampus Right} & $25$ & $53$ & $47$ \\
    \textbf{Amygdala Right} & $26$ & $54$ & $31$ \\
    \textbf{Ventral DC Right} & $27$ & $60$ & $61$ \\
    \hline
  \end{tabular}
  \label{tab:label_remap}
\end{table*}

\section{References}

\bibliographystyle{elsarticle-harv}
\bibliography{papers}

\end{document}